\documentclass{article}
\usepackage[table]{xcolor}
\usepackage{amssymb}
\PassOptionsToPackage{numbers, compress}{natbib}

\usepackage[eandd, final]{neurips_2026}
\usepackage{tikz}
\usepackage{wrapfig,lipsum}
\usepackage{comment}
\usetikzlibrary{arrows.meta}  %
\usetikzlibrary{positioning, arrows.meta, calc}
\usetikzlibrary{shapes.geometric, arrows.meta, positioning, fit, backgrounds}
\tikzset{arrow/.style={-{Stealth}, thick}}
\usepackage{booktabs}
\usepackage{enumitem}

\usepackage{listings}
\usepackage{parskip}
\usepackage[expansion=false]{microtype}
 
\definecolor{primary}{HTML}{1F4E79}
\definecolor{rulegray}{HTML}{4A4A4A}
\definecolor{codebg}{HTML}{F6F7F9}
\definecolor{codeframe}{HTML}{D0D4DA}
 
\usepackage{placeins}

\usepackage[utf8]{inputenc} 
\usepackage[T1]{fontenc}    
\usepackage{hyperref}       
\hypersetup{pdfborder={0 0 0}}
\usepackage{url}            
\usepackage{booktabs}       
\usepackage{amsfonts}       
\usepackage{nicefrac}       
\usepackage{microtype}      
\usepackage{graphicx}
\usepackage{subcaption}
\usepackage[most]{tcolorbox}

\tcbuselibrary{skins,breakable}
\usepackage{adjustbox}
\usepackage{listings}
\usepackage[expansion=false]{microtype}
\usepackage{parskip}
\usepackage{fontawesome5}

\usepackage{amsmath}

\let\plainclubsuit\clubsuit 

\newcommand{\cardfan}{%
  \tikz[
    baseline=0pt,
    x=0.8em, y=0.8em,
    line width=0.055em,
    line join=round
  ]{%
    \foreach \angle in {25,0,-25}{
      \draw[
        rotate=\angle,
        fill=white,
        rounded corners=0.04em
      ] (-0.28,0) rectangle (0.28,0.85);
    }
    \begin{scope}[rotate=-25]
      \node[inner sep=0pt, scale=0.35, transform shape]
        at (0,0.43) {$\plainclubsuit$};
    \end{scope}
  }%
}

\renewcommand{\clubsuit}{\text{\cardfan}}

\lstdefinelanguage{json}{
    basicstyle=\ttfamily\footnotesize,
    breaklines=true,
    columns=fullflexible,
    showstringspaces=false,
    string=[s]{"}{"},
    stringstyle=\color{blue!60!black},
    comment=[l]{:},
    commentstyle=\color{black},
    morestring=[b]",
}

\lstdefinestyle{prompt}{
  basicstyle=\ttfamily\footnotesize,
  backgroundcolor=\color{codebg},
  frame=single,
  rulecolor=\color{codeframe},
  framesep=5pt,
  xleftmargin=6pt,
  xrightmargin=6pt,
  breaklines=true,
  showstringspaces=false,
  columns=fullflexible,
  keepspaces=true,
  aboveskip=6pt,
  belowskip=6pt,
}

\newtcolorbox{promptbox}{
  enhanced, breakable,
  colback=white, colframe=primary!70,
  boxrule=0.5pt, arc=2pt,
  left=10pt, right=10pt, top=10pt, bottom=10pt,
}

\title{%
  \vspace{-0.35cm}%
  \begin{minipage}[c]{1.5cm}
    \includegraphics[width=\linewidth]{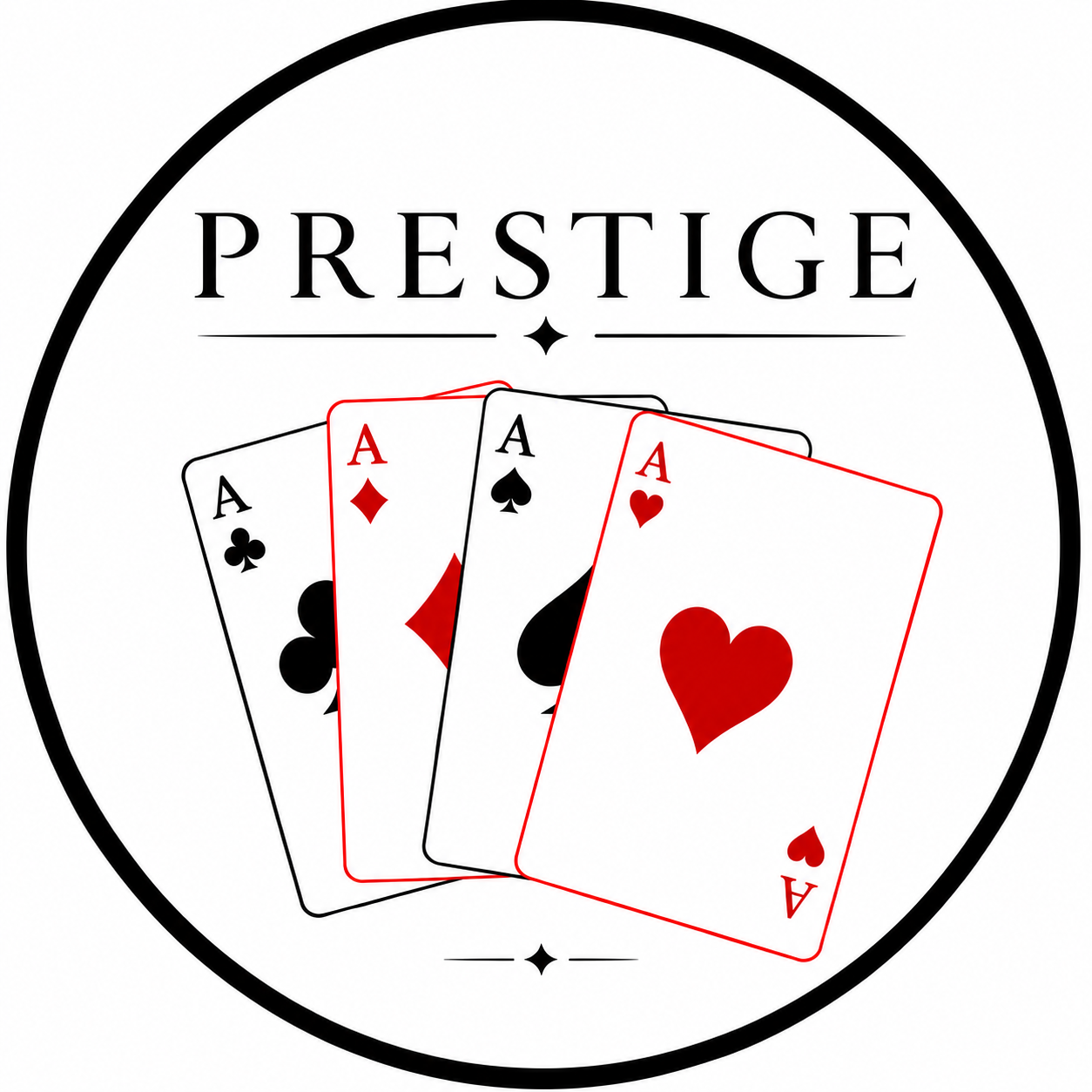}%
  \end{minipage}%
  \hspace{0.1cm}%
  \begin{minipage}[c]{\dimexpr\textwidth-1.6cm\relax}
    \centering
    The Prestige: Benchmarking Cognitive Visual\\
    Reasoning using Magic Tricks%
  \end{minipage}%
  \par\vspace{-0.35cm}%
}

\author{%
  Shuo Wen$^{1,2}$\thanks{Correspondence to: swen@utexas.edu} \quad
  Beixi Du$^{3,\clubsuit}$ \quad
  Edwin Meriaux$^{1,2,4,\clubsuit}$ \quad
  Junming Shi$^{1,2,\clubsuit}$ \quad
  Chloe Si$^{1,\clubsuit}$ \\
  \bfseries David Meger$^{1,2}$ \quad
  Doina Precup$^{1,2}$ \quad
  Gregory Dudek$^{1,2}$\\[0.6em]
  \normalfont\small
  $^{1}$McGill University \quad
  $^{2}$Mila -- Quebec AI Institute \quad
  $^{3}$Cornell University \\[0em]
  $^{4}$Universit\'e Paris-Saclay, CentraleSupélec \\
  $\clubsuit$ Equal contribution, listed alphabetically by last name.
}

\newtcolorbox{qbox}[1][]{
  colback=blue!5!white,
  colframe=blue!40!black,
  boxrule=0.5pt,
  arc=3pt,
  left=4pt, right=4pt, top=4pt, bottom=4pt,
  #1
}

\begin{document}

\maketitle

\begin{figure}[htbp]
    \centering
    \includegraphics[width=1\textwidth]{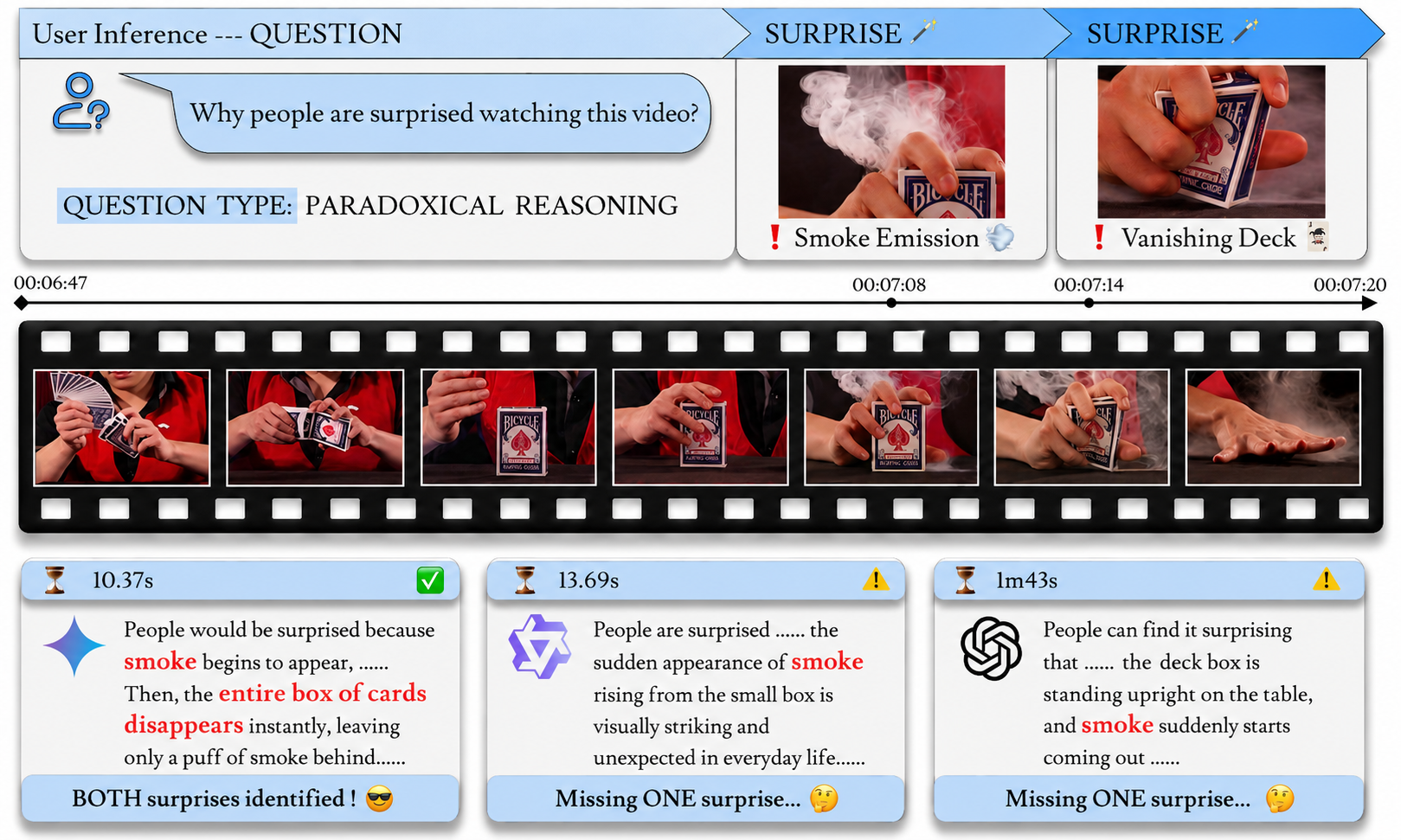}
    \caption{An example prompt to challenge representative VLMs to explain the elements of surprise in a video sequence. Results highlight current limitations in VLM visual reasoning. All models are evaluated using their latest web versions at the time of testing. Responses are excerpted, and images are cropped for brevity. \href{https://stevvwen.github.io/prestige-website/}{\textcolor{blue}{Project page}}}
    \label{fig:starting_img}
\end{figure}

\begin{abstract}
We introduce The Prestige, a video understanding benchmark that probes cognitive visual reasoning through professional magic performances, where deliberate misdirection challenges models to track events and explain unexpected outcomes. The benchmark comprises open-ended questions ranging from temporal and causal understanding to robustness against adversarial questions. Evaluation of leading open-source and proprietary VLMs reveals a substantial gap between even the strongest model and human performance. We also find systematic discrepancies between human and automated grading and identify recurring model failure modes. Ablations further show that models can select correct multiple-choice answers even when they fail to generate correct explanations independently, motivating further investigations into question formats. Together, these findings highlight limitations in both VLM reasoning and its evaluation, underscoring the need for more reliable video understanding models and more rigorous benchmarks.

\end{abstract}

\section{Introduction}
The illusion of magic fundamentally exploits the mechanisms of human commonsense reasoning. When a magician pulls a dove from a seemingly empty hat, our surprise stems not from the lack of observation but from our active, unconscious anticipation of logical outcomes. Experiencing this deception relies on what we define as \emph{cognitive flow}: a compositional process that involves tracking long-context visual cues, focusing attention on relevant details, and constructing predictive mental models without explicit language. Since magic deliberately subverts these deeply ingrained, multi-step reasoning processes, it presents a valuable domain for evaluating how intelligent systems parse visual narratives, model expectations, and react to counter-intuitive realities.

Consequently, professional magic performances offer unique, structural advantages for benchmarking the video understanding capabilities of Vision-Language Models (VLMs). First, many of the magic performances are engineered to be comprehended through observation alone, allowing benchmarks to isolate pure visual reasoning without reliance on language guidance or captions. Second, these performances naturally demand long-horizon memory and causal state tracking, where magicians frequently foreshadow elements that are revealed much later, deliberately weaving intricate subplots to misdirect the focus of the audience. Finally, because the core mechanics of advanced sleights are closely guarded secrets within the magic community, it is highly unlikely that models can rely on memorized training data to decipher the tricks.

The rapid development of VLMs~\citep{ bai2025qwen3,bai2025qwen25vl,chen2024longvila,deepmind2025gemini31pro,lin2024video,singh2025openai,zhu2025internvl3} has pushed video understanding toward stronger temporal modeling and longer-context reasoning. This trajectory motivates benchmarks to go beyond generic video comprehension to evaluate visual reasoning over temporally distant dependencies, multi-step reasoning, and causal visual clues. Although recent benchmarks have made important progress~\citep{chen2024cg,cheng2025video,li2022representation,xie2024funqa,yu2025vrbench}, many still rely on multiple-choice formats~\citep{fu2026video,wang2025lvbench,wu2024longvideobench,zhu2025mmr}. Such formats encourage reasoning shortcuts and only provide limited evidence that models can perform the cognitive visual reasoning required to track hidden state changes, maintain long-horizon expectations, and explain visually surprising outcomes. Our contributions are as follows:
\begin{itemize}
    \item We introduce \emph{The Prestige}, a dataset of professional magic performances designed to evaluate a diverse set of visual reasoning capabilities of VLMs, from short-term perception to complex cognitive-flow reasoning. To comprehensively test these skills, we define six distinct question categories covering a broad range of model abilities, yielding 2,177 open-ended questions grounded in visually sufficient, high-quality magic performances.

    \item Through extensive evaluations, we demonstrate a substantial performance gap between current state-of-the-art models and human baselines, establishing a new frontier for evaluating the cognitive visual reasoning abilities of VLMs. Our error analysis further reveals that current VLMs are primarily bottlenecked by limited cognitive flow, weak long-context reference, and unreliable visual identification.

    \item In addition, we conduct an ablation study demonstrating that multiple-choice formats artificially inflate model performance. This finding advocates for moving toward strictly open-ended question formats to ensure a more rigorous and unbiased evaluation of true visual reasoning abilities.
\end{itemize}

The full dataset can be found at: \faGithub\ \textcolor{blue}{\url{https://github.com/stevvwen/The_prestige}}

\section{Related Work}

\begin{table}[htbp]
    \centering
    \resizebox{\textwidth}{!}{%
    \begin{tabular}{l c c l c c c c c}
        \toprule
        \textbf{Dataset} & \textbf{\# videos}  & \textbf{\# QA} & \textbf{Format} & \textbf{Commonsense} & \textbf{Cognitive Flow} & \textbf{Multi-step} & \textbf{Long-context} & \textbf{Annotation} \\
        \midrule
        FunQA \citep{xie2024funqa}    & 4,365  & 311,950   & QA & $\checkmark$ & $\triangle$ & $\times$     & $\times$     & Manual+LLM \\
        Video-MME-v2 \citep{fu2026video}    & 800    & 3,200  & 8-way MC      & $\checkmark$     & $\times$ & $\checkmark$ & $\checkmark$ & Manual+LLM     \\
        Causal-VidQA \citep{li2022representation}    & 26,900   & 107,600   & MC   & $\checkmark$ & $\times$ & $\times$ & $\times$     & Manual     \\
        VRBench \citep{yu2025vrbench}    & 960   & 8,243   & MC\&QA   & $\times$ & $\times$ & $\checkmark$ & $\checkmark$     & Manual+LLM     \\
        Video-Holmes \citep{cheng2025video}    & 270   & 1,837   & MC   & $\checkmark$ & $\times$ & $\checkmark$ & $\times$     & Manual+LLM     \\

        CG-Bench \citep{chen2024cg}    & 1,219    & 12,129   & MC\&QA   & $\times$ & $\times$ & $\checkmark$ & $\checkmark$     & Manual     \\

        MMR-V \citep{zhu2025mmr}    & 317   & 1,257   & MC   & $\checkmark$ & $\times$ & $\checkmark$ & $\checkmark$     & Manual     \\
        
        LongVideoBench \citep{wu2024longvideobench}    & 3,763   & 6,678   & MC   & $\times$ & $\times$ & $\times$ & $\checkmark$     & Manual     \\
        LVBench \citep{wang2025lvbench}    & 103    & 1,549   & MC & $\times$ & $\times$ & $\checkmark$     & $\checkmark$ & Manual+LLM \\
        \midrule
        \textbf{Ours}    & \textbf{180} & \textbf{2,177} & \textbf{QA} & $\pmb \checkmark$ & $\pmb \checkmark$ & $\pmb \checkmark$ & $\pmb \checkmark$ & Manual \\
        \bottomrule
    \end{tabular}
    }
    \caption{Comparison of different video understanding benchmarks. \textbf{MC} stands for multiple choice questions, \textbf{QA} stands for open Question-Answering. \textbf{Cognitive Flow} denotes whether the benchmark centrally evaluates reasoning over videos designed to examine the cognitive visual reasoning of the models. $\triangle$ indicates partial coverage.}
    \label{tab:comparison}
\end{table}

\subsection{VLMs for video understanding}
VLMs for videos have rapidly evolved from short-video conversational systems toward stronger temporal modeling and longer-context understanding. Early advancements like Video-LLaVA~\citep{lin2024video}, VideoLLaMA-2~\citep{cheng2024videollama}, and LLaVA-Video~\citep{zhang2024llava} establish strong chat-style interfaces for unified image and video understanding. VideoLLaMA 3~\citep{zhang2025videollama} further strengthened this line with a more vision-centric design for both image and video understanding. More recently, models such as LongVILA~\citep{chen2024longvila},
Qwen2.5-VL~\citep{bai2025qwen25vl}, Qwen3VL~\citep{bai2025qwen3},
InternVL3~\citep{zhu2025internvl3}, GLM-4.6V~\citep{hong2025glm},
InternVL3.5~\citep{wang2025internvl3}, and MiniCPM-V
4.5~\citep{yu2025minicpm} have advanced multimodal perception and reasoning, specifically enhancing general-purpose capabilities and long-video comprehension. Further, InternVideo2.5~\citep{wang2025internvideo2}, LongVU~\citep{shen2024longvu}, and LongVILA-R1~\citep{chen2025scaling} emphasize long-context modeling, efficient compression, and reasoning over
extended video streams. In parallel, proprietary systems such as GPT-5~\citep{singh2025openai} and Gemini3.1-Pro~\citep{deepmind2025gemini31pro} have further advanced frontier multimodal reasoning capabilities.

\subsection{Benchmarks for video understanding}
Benchmarks for video understanding have progressed from temporal perception
and general QA toward long-context and reasoning-centric evaluation.
MVBench~\citep{li2024mvbench} and TempCompass~\citep{liu2024tempcompass}
evaluate fine-grained temporal understanding, including motion, order, and
direction. Video-MME~\citep{fu2025video} broadens evaluation across video
types, durations, and auxiliary modalities, while MLVU~\citep{zhou2025mlvu}, LVBench~\citep{wang2025lvbench}, LongVideoBench~\citep{wu2024longvideobench}, and ReXTime~\citep{chen2024rextime} extend evaluation to longer videos and
temporally separated evidence. Video-MME-v2~\citep{fu2026video} further
introduces a hierarchical evaluation of video understanding, including
counterintuitive comprehension under physical-world reasoning. In parallel, CG-Bench evaluates clue-grounded long-video question answering~\citep{chen2024cg}, Video-Holmes studies complex reasoning over suspense short films~\citep{cheng2025video}, MMR-V focuses on long-range multi-frame evidence and hidden information~\citep{zhu2025mmr}, and VRBench evaluates reasoning in long narrative videos with timestamped intermediate steps~\citep{yu2025vrbench}. Causal-VidQA~\citep{li2022representation} studies explanation, prediction, and counterfactual reasoning in video. These settings require models to aggregate evidence over time, but they do not examine the deeper cognitive flow of the model. The closest prior benchmark to our setting is FunQA~\citep{xie2024funqa}, which includes MagicQA as part of a broader benchmark on counterintuitive videos. However, FunQA mixes magic with humor, creative performances, visual-illusion-style effects, and mainly treats magic as a local anomaly, whereas our benchmark focuses on staged magic routines in which the performer deliberately controls attention, hides state changes, and delays the reveal. We discuss why we choose magic as the testbed in Appendix~\ref{app:why_magic}. The qualitative comparison between datasets can be found in Table~\ref{tab:comparison}.



\section{Methodology}
\label{sec:methodology}


\subsection{Video Collection}

To control benchmark quality, we select videos that meet the Video Quality Standards defined in Appendix~\ref {sec:vqc}. We ensure the videos are professional, ethical, and high-resolution. The videos must be entirely comprehensible through visual cues alone, independent of audio or text, and not captioned. Additionally, the performances must demonstrate intentional causal coherence and high contextual density. This process is fully introduced in Appendix~\ref{sec:sourcing}.

\subsection{Question and Answer generation}
\label{sec:qa-generation}
\begin{figure}
    \centering\includegraphics[width=1\linewidth]{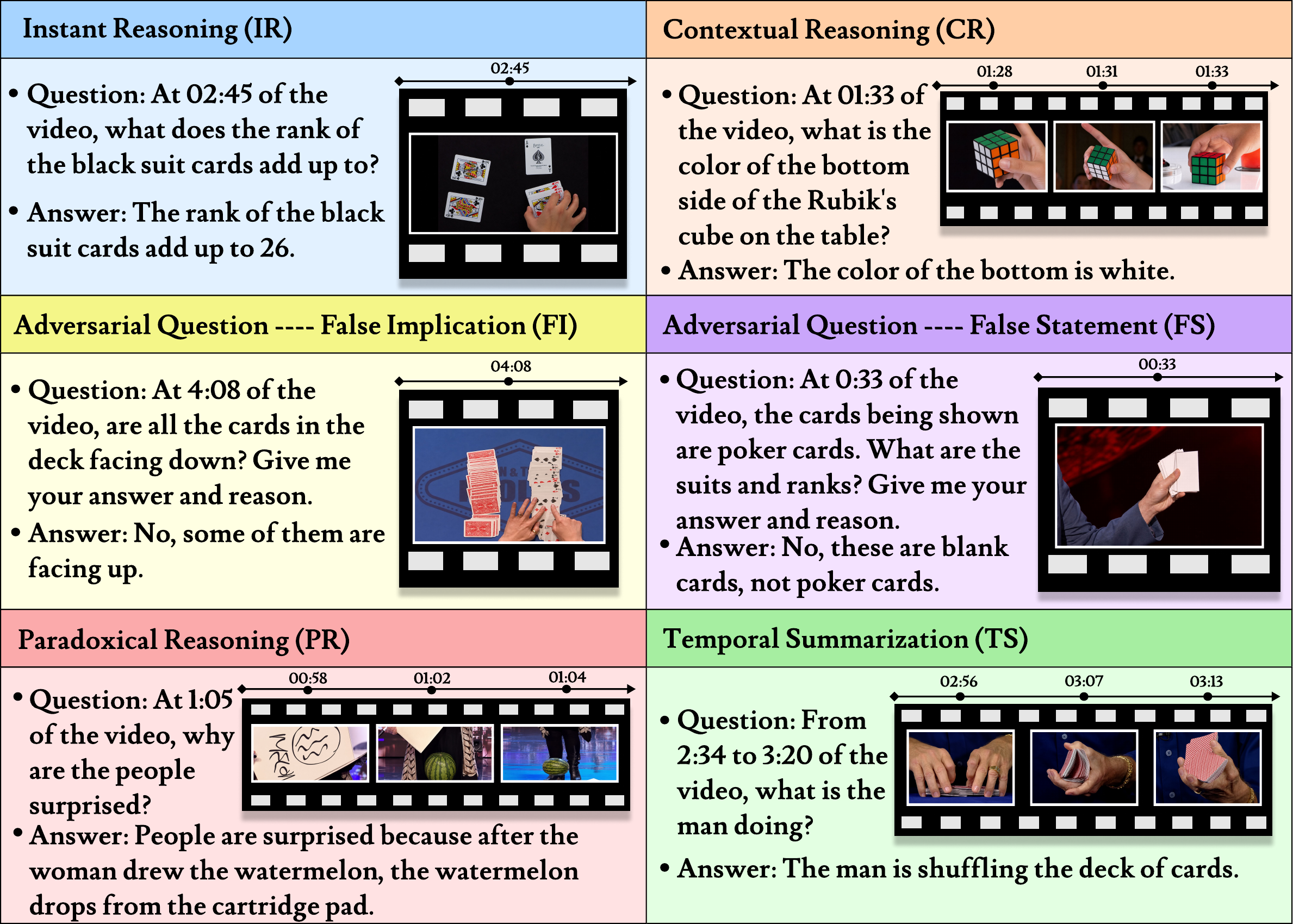}
    \caption{Examples of the six distinct categories of video reasoning tasks evaluated: Instant Reasoning (IR), Contextual Reasoning (CR), Temporal Summarization (TS), Adversarial Questions—False Implication (AQ.FI) and False Statement (AQ.FS), and Paradoxical Reasoning (PR). Each panel displays a sample question, the ground-truth human answer, and a film strip of the relevant video clip. Images are zoomed in for clarity. Representative frames are only selected for demonstration purposes. In actual evaluations, models receive frames sampled uniformly at 1 FPS over the available temporal interval, up to a maximum duration of four minutes.}
    \label{fig:sampleimage}
\end{figure}

Following the video selection, we systematically construct Question-Answer pairs to comprehensively evaluate diverse aspects of video understanding. To effectively probe the limits of VLM reasoning, we define a taxonomy of core question categories, each targeting distinct aspects of visual, temporal, and commonsense reasoning in multi-step and long-dependency settings. Below are the six categories of questions generated for this dataset. A more in-depth introduction to each category can be found in Appendix~\ref{sec:question-types}, and the generation pipeline is described in Appendix~\ref{AP}. Visual examples are provided in Figure~\ref{fig:sampleimage}.

\textbf{Instant Reasoning (IR): } These questions aim to assess the instantaneous reasoning abilities of VLMs, based on the visual context of the current frame. Such frame-level analysis forms the necessary foundation for complex temporal understanding and reasoning chains.

\textbf{Contextual Reasoning (CR): } These questions examine the memory and relational reasoning abilities of VLMs. The answers to such questions require context not present in the frame being shown. This context must be understood and constructed by the model from the preceding frames.

\textbf{Paradoxical Reasoning (PR):} These questions are designed to require the models to explain the counterintuitive nature of magic tricks. During a magic trick, certain parts of the act are explicitly set up to surprise the audience. Understanding why these scenes are surprising is important as it requires VLMs to perform multi-step reasoning by tracking object states, predicting expected outcomes using common sense, and detecting deviations from observed reality.

\textbf{Adversarial Questions (AQ):} These questions introduce premises or implications that contradict visual evidence or memory, requiring VLMs to reject false information and ground their reasoning in observations. \textbf{False Implication (AQ.FI)} questions suggest a false proposition through their wording without explicitly asserting it as fact. In contrast, \textbf{False Statement (AQ.FS)} questions explicitly make a false claim and construct a follow-up question that assumes the premise is true. A correct response must challenge that premise and provide a correction grounded in visual evidence.

\textbf{Temporal Summarization (TS):} These questions require VLMs to write a compact, high-level summary of events from a given time frame. This does not examine an enumeration of low-level visual details. Models must capture the core actions, avoiding both unnecessary specifics and overly broad generalizations. This category isolates the model's distinct ability to compress long temporal windows into key takeaways.

\subsection{Dataset evaluation}

Once these questions are generated, all the questions are peer reviewed to ensure they meet the Question-Answer Quality Standard (QAQS). The QAQS is described in Appendix~\ref{sec:vqc}, and the review process in Appendix~\ref{Peer_review}.

\subsection{Statistics of the dataset}

\begin{wrapfigure}{r}{0.40\textwidth}
    \centering
    \includegraphics[width=\linewidth]{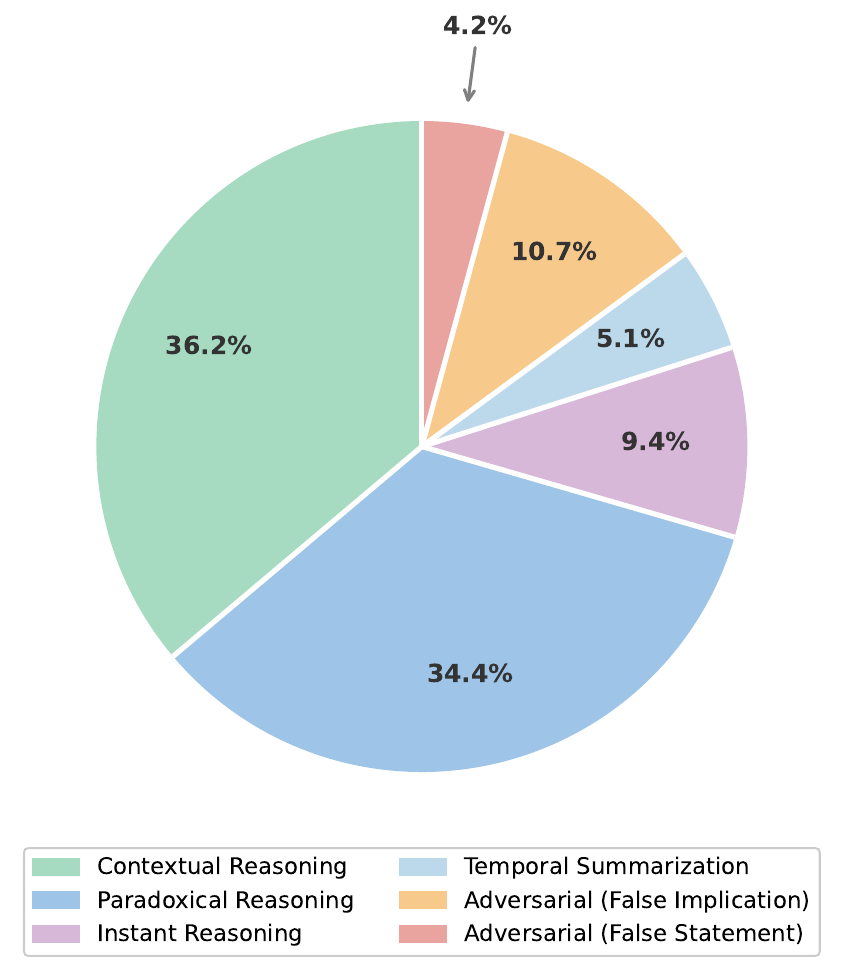}
    \caption{Question Type Distribution}
    \label{fig:question_type_distribution}
\end{wrapfigure}

The final benchmark contains 180 videos and 2,177 open-ended questions, drawn from unique routines performed by over 150 magicians. Figure~\ref{fig:question_type_distribution} displays the distribution of question types. The benchmark covers a broad spectrum of cognitive visual reasoning skills, with Contextual Reasoning (36.2\%), Paradoxical Reasoning (34.4\%), Instant Reasoning (9.4\%), False Implication (10.7\%), False Statement (4.2\%), and Temporal Summarization (5.1\%). Beyond question-type diversity, the keyword vocabulary in Figure~\ref{fig:word_cloud} shows that the questions span varied props and effects, preventing the benchmark from collapsing into a narrow subset of card tricks or object-recognition problems. More comments about the dataset size and question type distribution can be found in Appendix~\ref{sec:datasetcomments}.

\section{Experiments}

We evaluate a diverse set of VLMs that natively support long-video or multi-frame image understanding. On the open-source model side, we choose a comprehensive suite of recent representative VLMs: Qwen3VL~\citep{bai2025qwen3}, Qwen3.6~\citep{qwen36_35b_a3b}, InternVideo2.5~\citep{wang2025internvideo2}, InternVL3.5~\citep{wang2025internvl3}, MOSS-VL-Instruct~\citep{mossvl}, Cambrian-S~\citep{yang2026cambrian}, Keye-VL-1.5~\citep{yang2025kwai}, MiMo-VL~\citep{yue2025mimo}, and Kimi-K2.6~\citep{moonshot2026kimik26}. In addition, we include leading proprietary models to establish the current best performance frontiers, including Gemini3.1-Pro~\citep{deepmind2025gemini31pro} and Flash-Lite~\citep{googledeepmind2026gemini31flashlite}, GPT-5.2~\citep{openai2025gpt52}, and Qwen3.6-Plus~\citep{qwenteam2026qwen36plus}. For consistent comparison under API constraints, all models receive video frames sampled at 1 FPS from a window extending up to four minutes backward from each question’s timestamp. To establish the human baseline, we conducted an independent cross-review protocol, where each question was answered by a reviewer other than its author under the same evaluation protocol as the models. Incorrect answers were recorded to compute human accuracy rates, providing a reference for model performance relative to human visual understanding and reasoning.

We run all open-source models on A100 GPU clusters, excluding the Qwen family and Kimi-K2.6, which we accessed through the official API. For all closed-source models, we utilized their officially provided APIs.

\subsection{Benchmark Results}

\begin{table*}[htbp]
    \centering
    \resizebox{\textwidth}{!}{
        \begin{tabular}{p{4.7cm} | c c c c c c c}
        \toprule
        \textbf{Model} \hfill \textbf{Size} & \textbf{IR} & \textbf{CR} & \textbf{PR} & \textbf{AQ.FI} & \textbf{AQ.FS} & \textbf{TS} & \textbf{Overall} \\
        \midrule

        \textit{\qquad Open-source models} & & & & & & & \\
        \midrule
        Qwen3VL-Thinking \citep{bai2025qwen3} \hfill 8B & 40.00 & 34.69 & 9.76 & 24.46 & 15.22 & 17.86 & 23.84 \\
        Qwen3VL-Instruct \citep{bai2025qwen3} \hfill 8B & 46.34 & 39.90 & 17.25 & 18.88 & 20.65 & 29.46 & 29.12 \\
        InternVideo2.5-Chat \citep{wang2025internvideo2} \hfill 8B & 30.24 & 27.57 & 3.74 & 3.86 & 5.43 & 8.93 & 15.20 \\
        InternVL3.5 \citep{wang2025internvl3} \hfill 8B & 25.37 & 26.81 & 4.14 & 19.31 & 9.78 & 12.50 & 16.63 \\
        MiMo-VL-SFT-2508 \citep{yue2025mimo} \hfill 7B & 22.44 & 22.11 & 1.87 & 21.89 & 15.22 & 8.04 & 14.15 \\
        MOSS-VL-Instruct \citep{mossvl} \hfill 11B & 36.59 & 33.16 & 12.43 & 30.47 & 7.61 & 28.57 & 24.76 \\
        Qwen3.6 \citep{qwen36_35b_a3b} \hfill 35A3B & 51.71 & 48.16 & 28.34 & 31.76 & 21.74 & 37.50 & 38.26 \\
        Cambrian-S \citep{yang2026cambrian} \hfill 7B & 24.88 & 22.36 & 4.55 & 20.17 & 8.70 & 10.71 & 15.07 \\
        Keye-VL-1.5 \citep{yang2025kwai} \hfill 8B & 30.73 & 27.83 & 5.61 & 23.18 & 14.13 & 10.71 & 18.51 \\
        
        Kimi-K2.6 \citep{moonshot2026kimik26} \hfill 1.1TA32B & 54.15 & 47.14 & 31.28 & 30.47 & 27.17 & 40.18 & 39.37 \\
        \midrule
        
        \textit{\qquad Proprietary models} & & & & & & & \\
        \midrule
        GPT-5.2 \citep{openai2025gpt52} \hfill UKN & 53.17 & 46.12 & 29.41 & 30.47 & 15.22 & 38.39 & 37.67 \\
        Gemini3.1-Flash-Lite \citep{googledeepmind2026gemini31flashlite} \hfill UKN & 55.12 & 49.94 & 34.89 & 41.63 & 18.48 & 41.96 & 42.63 \\
        Qwen3.6-Plus \citep{qwenteam2026qwen36plus} \hfill UKN & 49.76 & 51.46 & 35.16 & 28.33 & 16.30 & 39.29 & 41.11 \\
        Gemini3.1-Pro \citep{deepmind2025gemini31pro} \hfill UKN & 52.68 & 50.44 & 43.32 & 45.06 & 55.43 & 35.71 & 47.08 \\
        \midrule
        \rowcolor{gray!25}
        Human baseline \hfill NA & 93.17 & 93.01 & 94.39 & 91.42 & 83.70 & 90.18 & 92.79 \\
        
        \bottomrule
    \end{tabular}
    } 

    \caption{Model accuracy by question type across 14 models over the full dataset. All model results are scored by human evaluation.}
    \label{tab:experiments}
\end{table*}

\textbf{Human baseline Interpretation.}
The human baseline achieves an overall accuracy of 92.79\%, indicating that our dataset is challenging but reliably answerable by careful human viewers. Human performance remains consistently high across question types, with the highest score on PR (94.39\%), followed by IR (93.17\%) and CR (93.01\%). The peak PR score supports our dataset design: humans can reliably solve the central cognitive-flow challenge of the benchmark by tracking the setup, forming an expected trajectory, and recognizing how the reveal contradicts that expectation. This indicates that PR questions reflect visually grounded consensus rather than subjective ``surprise'' judgments.

The lowest human scores are on AQ.FS (83.70\%), suggesting that adversarial questions with explicit false premises are difficult even for humans. AQ.FI (91.42\%) is higher, indicating that humans are more robust to misleading assumptions compared to false statements. TS also shows a slightly lower score (90.18\%), reflecting the variability of open-ended summaries and the difficulty of compressing a temporal window into a concise high-level description without losing key details. Overall, the human baseline highlights both the reliability and difficulty of our benchmark, with the hardest cases involving the same cognitive reasoning abilities we aim to test in VLMs.

\begin{wrapfigure}{hr}{0.40\textwidth}
    \centering
    \includegraphics[width=\linewidth]{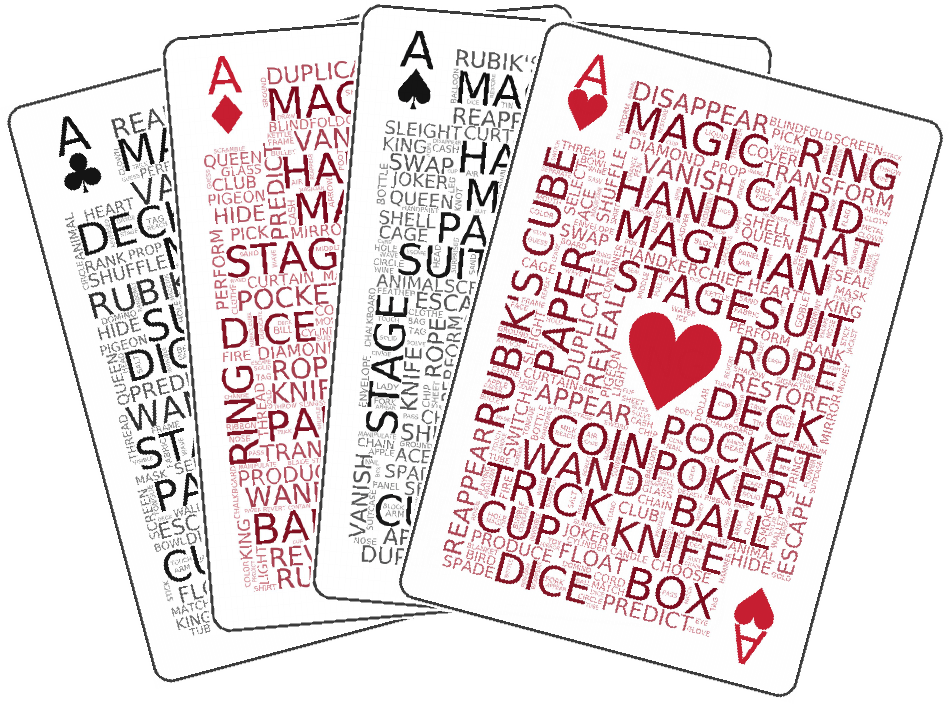}
    \caption{Word cloud}
    \label{fig:word_cloud}
\end{wrapfigure}

\textbf{Model Performance Analysis.}
Table~\ref{tab:experiments} reports human-graded accuracy for all 14 models. Gemini3.1-Pro performs best overall at 47.08\%, but remains 45.71 percentage points below the human baseline (92.79\%). Moreover, no model leads every category. Gemini3.1-Pro achieves the highest PR (43.32\%), AQ.FI (45.06\%), and AQ.FS (55.43\%) scores; Gemini3.1-Flash-Lite leads IR (55.12\%) and TS (41.96\%); and Qwen3.6-Plus leads CR (51.46\%). These complementary strengths indicate that current models have not yet developed uniformly strong visual reasoning across the benchmark's perceptual, temporal, contextual, and adversarial demands.

The strongest open-source models, Kimi-K2.6 (39.37\% overall) and Qwen3.6 (38.26\%), both outperform GPT-5.2 (37.67\%) and substantially exceed the smaller open-source models, whose overall scores range from 14.15\% to 29.12\%. Nevertheless, even the best model remains well below humans in every category, with particularly large deficits on PR and TS. The results therefore show meaningful progress among larger models while preserving substantial headroom in long-horizon cognitive reasoning and temporal understanding.

\subsection{Comparison of Human and Machine Grading}

We then compare the grading between humans and machines to give a comprehensive view of the open QA evaluation landscape. For reproducibility, machine grades are determined by majority voting among Qwen3.7-plus-2026-05-26, Gemini-3.5-flash-lite, and GPT-5.6-Luna, accessed through their official APIs with temperature 0.0 and seed 42. Each grader independently evaluates every model response coupled with the corresponding video frames for reference, using the grading prompt in Figure~\ref{fig:grading-prompt}. The majority voting determines the final binary grade for each QA pair. Full machine-grading results are reported in Appendix~\ref{sec:machine_grading}.

Table~\ref{tab:human_machine_difference} compares human and machine grading across all 14 models. Entries are human accuracy minus machine accuracy, in percentage points and the final row summarizes the mean and standard deviation across models.

\begin{table*}[htbp]
\centering
\setlength{\tabcolsep}{4pt}
\resizebox{\textwidth}{!}{%
\begin{tabular}{@{}l|ccccccc@{}}
\toprule
\textbf{Model} & \textbf{IR} & \textbf{CR} & \textbf{PR} & \textbf{AQ.FI} & \textbf{AQ.FS} & \textbf{TS} & \textbf{Overall} \\
\midrule
\multicolumn{8}{l}{\textit{\qquad Open-source models}} \\
\midrule
Qwen3VL-Thinking-8B    & 7.80 & 3.81 & 6.42 & 3.86  & 10.87 & 8.04  & 5.60 \\
Qwen3VL-Instruct-8B    & 8.78 & 5.47 & 8.96 & 4.29  & 9.78  & 13.39 & 7.44 \\
InternVideo2.5-Chat-8B & 6.83 & 4.32 & 2.54 & 0.86  & 4.34  & 7.14  & 3.72 \\
InternVL3.5-8B        & 7.81 & 3.18 & 2.94 & 3.86  & 3.26  & 10.71 & 4.00 \\
MiMo-VL-SFT-2508-7B   & 8.29 & 2.80 & 1.07 & 7.30  & 8.70  & 3.58  & 3.49 \\
MOSS-VL-Instruct-11B  & 4.39 & 2.66 & 5.21 & 5.58  & 6.52  & 16.96 & 4.92 \\
Qwen3.6-35A3B         & 6.83 & 4.96 & 9.22 & -2.57 & 6.52  & 11.61 & 6.20 \\
Cambrian-S-7B         & 6.83 & 3.17 & 3.21 & 8.15  & 8.70  & 6.25  & 4.46 \\
Keye-VL-1.5-8B        & 5.85 & 2.93 & 2.27 & 2.58  & 6.52  & 2.67  & 3.08 \\
Kimi-K2.6-1.1TA32B    & 1.95 & 1.90 & 6.95 & -3.44 & 1.08  & 12.50 & 3.59 \\
\midrule
\multicolumn{8}{l}{\textit{\qquad Proprietary models}} \\
\midrule
GPT-5.2              & 5.37 & 3.17 & 1.74 & 15.45 & 6.52  & -0.90 & 4.14 \\
Gemini3.1-Flash-Lite  & 7.80 & 3.69 & 5.61 & 3.86  & 1.09  & 6.25  & 4.78 \\
Qwen3.6-Plus         & 5.86 & 3.43 & 9.09 & -0.85 & 2.17  & 4.47  & 5.14 \\
Gemini3.1-Pro        & 2.44 & 2.03 & 5.62 & 3.86  & -1.09 & 0.89  & 3.30 \\
\midrule
AVG\,{\scriptsize \emph{SD}} & 6.20\,{\scriptsize \emph{2.08}} & 3.39\,{\scriptsize \emph{1.01}} & 5.06\,{\scriptsize \emph{2.83}} & 3.77\,{\scriptsize \emph{4.76}} & 5.36\,{\scriptsize \emph{3.63}} & 7.40\,{\scriptsize \emph{5.13}} & 4.56\,{\scriptsize \emph{1.23}} \\
\bottomrule
\end{tabular}%
}
\caption{Signed percentage-point difference between human and machine grading scores, computed as human score minus machine score. The final row reports the mean for each column, with the corresponding standard deviation.}
\label{tab:human_machine_difference}
\end{table*}

Human grading yields higher overall scores for every model, with a mean gap of 4.56 percentage points (SD 1.23), ranging from 3.08 for Keye-VL-1.5 to 7.44 for Qwen3VL-Instruct. The gap varies by question type: TS has the largest mean difference (7.40 points), while CR has the smallest (3.39 points). TS and AQ.FI also show the greatest variation across models (SD 5.13 and 4.76, respectively). Negative differences in several AQ.FI entries, Gemini3.1-Pro's AQ.FS score, and GPT-5.2's TS score show that machine grading is not uniformly stricter. These aggregate differences indicate sensitivity to both the model and question type, but do not establish agreement on individual answers.

\subsection{Error Analysis}
To better understand the qualitative failure modes of frontier VLMs on \emph{The Prestige}, we conduct an error analysis on representative model outputs. Specifically, we randomly sample 100 incorrectly answered questions from each of Gemini3.1-Pro~\citep{deepmind2025gemini31pro}, GPT-5.2~\citep{openai2025gpt52}, and Kimi-K2.6~\citep{moonshot2026kimik26}. Each sampled error is manually assigned to one of seven error categories.

\textbf{Incorrect Visual Identification (IVI).}
IVI includes errors caused by incorrect recognition of evidence directly visible at the queried timestamp, such as misidentifying objects, misreading card ranks or suits, or miscounting items.

\textbf{Context Referral Deficiency (CRD).}
CRD captures failures in long-context reference and state tracking, where the model fails to retrieve information established earlier, such as a previously shown object identity or a hidden object location. These errors occur when the answer is not directly visible at the queried frame but can be inferred by linking the current scene to prior evidence.

\textbf{Lack of Cognitive Flow (LCF).}
LCF captures failures to follow the evolving causal structure of a magic routine. The model may recognize local visual events but fail to connect the setup, misdirection, and reveal, or may reuse an earlier surprise to explain a later, distinct one.

\textbf{Lack of Robustness (LR).}
LR describes failures to resist misleading or false premises, especially in adversarial questions. These errors occur when the model accepts an incorrect assumption in the questions instead of rejecting it and grounding the answer in visual evidence.

\textbf{Incorrect Output Formatting (IOF).}
IOF includes failures to follow the required response format, such as never-ending responses, refusals to answer, or missing required components.

\textbf{Human Intention Mismatch (HIM).}
HIM includes answers that are plausible or partially correct but fail to match the intended specificity or reasoning target of the question. Examples include giving generic explanations such as ``sleight of hand,'' over-detailing temporal summaries, or omitting explicitly requested reasoning.

\textbf{Hallucination (HAL).}
HAL refers to unsupported visual or causal claims, including invented objects, nonexistent actions, or incorrect surprises that substantially diverge from the ground-truth.

\begin{wrapfigure}{R}{0.40\textwidth}
\centering
\captionsetup{font=normalsize}

\includegraphics[width=\linewidth]{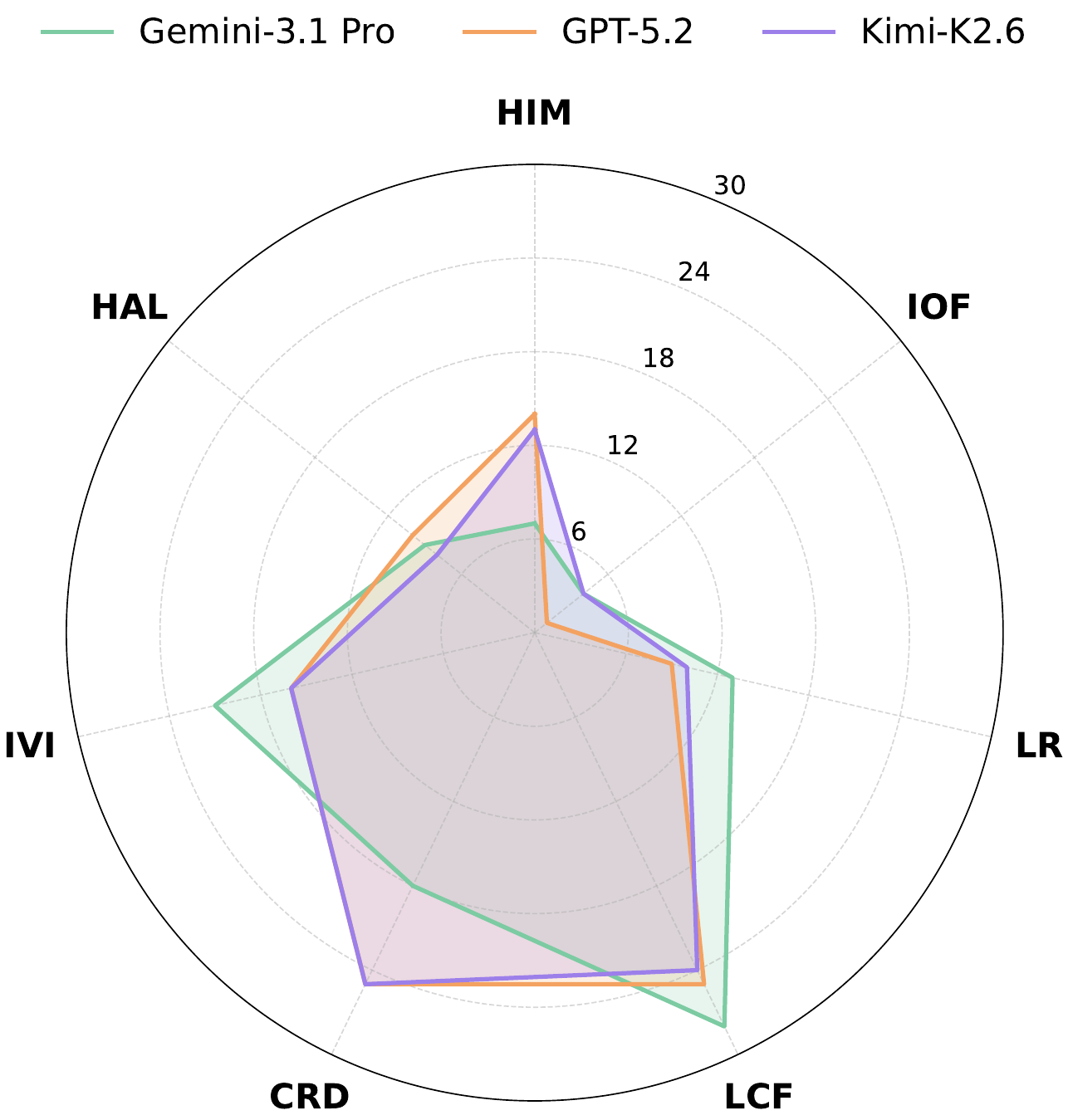}
\caption{Error Type Distribution}
\label{fig:error_analysis}

\vspace{6pt}

\includegraphics[width=\linewidth]{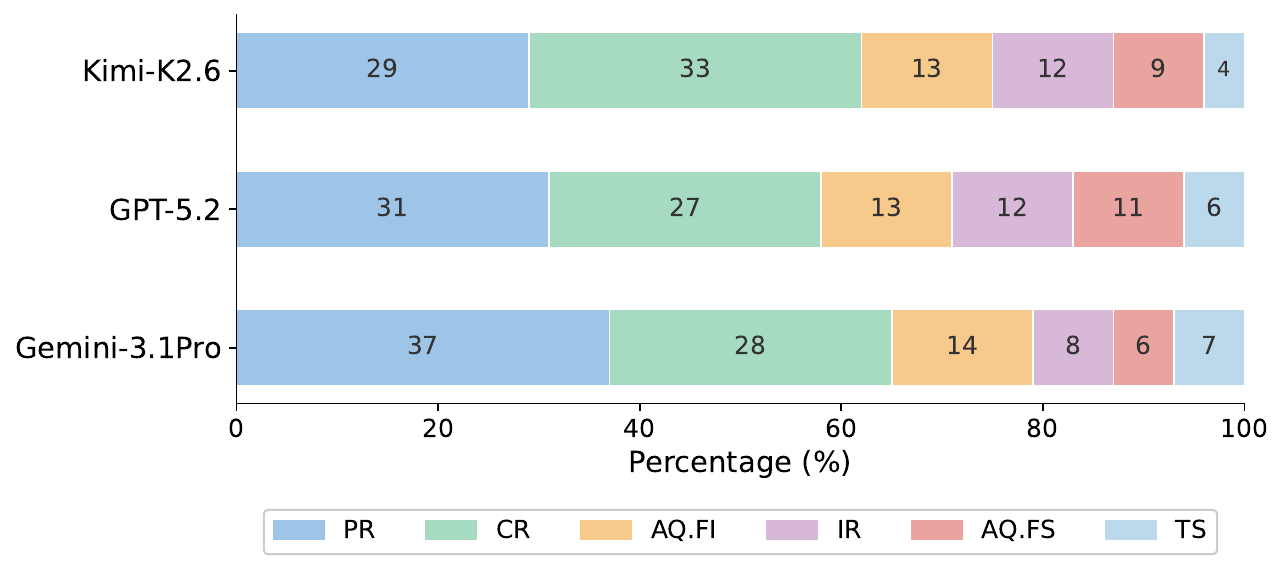}
\caption{Question Type Distribution of Sampled Errors}
\label{fig:error_question_type}

\vspace{-10pt}
\end{wrapfigure}

Figure~\ref{fig:error_analysis} shows that LCF, CRD, and IVI together account for 67\%, 66\%, and 65\% of sampled errors in Gemini3.1-Pro, GPT-5.2, and Kimi-K2.6, respectively, identifying cognitive-flow reasoning, long-context reference, and local visual identification as the dominant failure modes. HAL and IOF are less frequent, together accounting for 13\%, 11\%, and 12\%, respectively. Gemini3.1-Pro has the fewest sampled CRD errors and the highest CR accuracy among these models, consistent with stronger long-context reference ability.

These comparisons are descriptive. Each model contributes only 100 randomly sampled errors, and small differences may reflect sampling variation or borderline annotation decisions despite a shared rubric. The sampled question-type distributions are broadly comparable (Figure~\ref{fig:error_question_type}), but associations between question and error types may also affect the observed differences.

Together with Table~\ref{tab:experiments}, these findings highlight recurring difficulties in rejecting false premises and following a routine's cognitive flow: identifying the central surprise, linking it to the setup, and summarizing the essential steps.

\section{Ablations on the effects of MC Questions}

We evaluate the effect of question format on Qwen3.6-Plus, Kimi-K2.6, and Gemini3.1-Flash-Lite with two experiments: 200 questions randomly sampled from the benchmark without conditioning on open-ended correctness, and a separate diagnostic sample of 50 questions from human evaluation that all three models failed to answer in the open-ended setting. For both experiments, we compare open QA with 8-way and 4-way multiple-choice answering using the same question-conversion procedure. Following prior work~\citep{zhu2025mmr,cheng2025video}, we use an LLM to generate seven distractors for each question, designed to be semantically similar to the correct answer but factually incorrect. The 8-way setting includes the correct answer and all seven distractors. The 4-way setting includes the correct answer and three distractors randomly sampled from that pool. We randomly permute the option order to mitigate option-position bias and extract the selected answer using regular expression matching.

\begin{figure}[tbp]
    \centering
    \includegraphics[width=\textwidth]{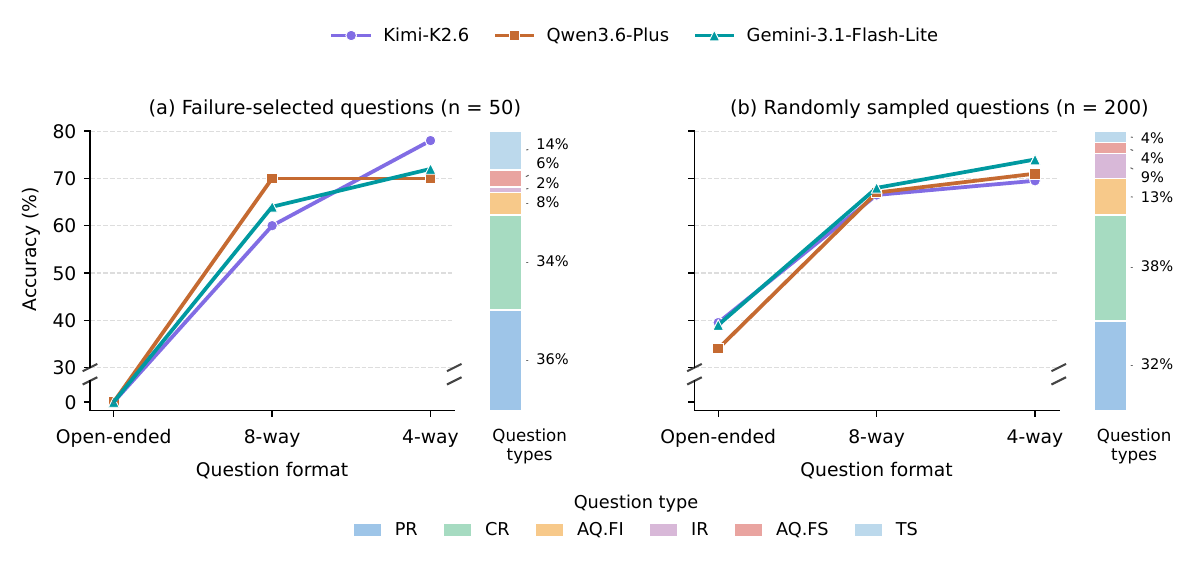}
    \caption{Ablation study on the effects of multiple-choice questions for (a) 50 failure-selected and (b) 200 randomly sampled questions. Stacked bars show the question-type distributions.}
    \label{fig:ablation}
\end{figure}

\textbf{Randomly sampled questions.} Figure~\ref{fig:ablation}(b) reports the comparison on 200 randomly sampled questions. Qwen3.6-Plus achieves 34.0\%, 67.0\%, and 71.0\% accuracy in the open-ended, 8-way, and 4-way settings, respectively. The corresponding scores are 39.0\%, 68.0\%, and 74.0\% for Gemini3.1-Flash-Lite, and 39.5\%, 66.5\%, and 69.5\% for Kimi-K2.6. Averaged across the three models, accuracy increases from 37.5\% for open-ended answering to 67.2\% for 8-way and 71.5\% for 4-way multiple choice, corresponding to gains of 29.7 and 34.0 percentage points, respectively. 

\textbf{Failure-selected questions.} Figure~\ref{fig:ablation}(a) shows an analysis of 50 questions that all three models failed in the open-ended setting. The open-ended accuracy is therefore 0\% by construction. In the 8-way setting, Qwen3.6-Plus, Kimi-K2.6, and Gemini3.1-Flash-Lite achieve 70\%, 60\%, and 64\%, respectively; their 4-way scores are 70\%, 78\%, and 72\%. The mean 4-way gain is 73.3 percentage points.

Together, these comparisons show that providing candidate answers substantially increases measured accuracy and can conceal failures to generate an adequate answer independently. These findings motivate further study of how question format shapes the assessment of model capabilities for tasks requiring qualitative reasoning, rather than the unambiguous numerical answers found in many mathematical problems.

\section{Discussion \& Conclusion}

In this paper, we introduce \emph{The Prestige}, a video understanding benchmark for testing cognitive visual reasoning of VLMs featuring professional magic performances, consisting of 2,177 open-ended questions that test short-term, multi-step, and long-range visual reasoning, as well as resistance to adversarial questions. Across 14 leading proprietary and open-source VLMs, the best model, Gemini3.1-Pro, scores 47.08\%, leaving a 45.71 percentage-point gap below the 92.79\% human baseline. Sampled error evaluation further reveals recurring failures in cognitive flow (LCF), context referral (CRD), and visual identification (IVI). Specifically, we find that VLMs consistently struggle when answering requires recalling an object absent from the current frame. They sometimes miss magic surprises that human reviewers commonly perceive and frequently fail to explicitly reject false information in adversarial settings. These failure modes reveal limited robustness in long-horizon tracking and multi-step reasoning within adversarial narratives.

Our comparison of human and majority voting machine grading shows that machine evaluation yields lower overall scores for all 14 models, with a mean difference of 4.56 percentage points and a standard deviation of 1.23 across models. The sign and magnitude of the differences vary by question category, with machine scores exceeding human scores in some model--category combinations. These aggregate differences indicate sensitivity to the grading protocol and motivate further improvement of automatic grading for open-ended visual reasoning tasks.

Furthermore, on 200 randomly sampled questions, mean accuracy across three models rises from 37.5\% with open-ended answers to 67.2\% with eight choices and 71.5\% with four choices. The 34.0-percentage-point gain in the 4-way setting shows that question format substantially affects measured performance beyond the failure-selected sample. The larger 73.3-percentage-point gain on 50 jointly failed questions further reveals the inflation effect. These comparisons demonstrate that selecting a supplied answer and generating an adequate answer independently are evaluation tasks with huge differences, supporting the diagnostic value of open-ended responses.

\textbf{Impact} We envision \emph{The Prestige} as a testbed for developing models that are better at visual reasoning. The questions and identified failure modes may inform future model training and more diagnostic evaluation protocols. Beyond evaluation, these capabilities may also be transferred to real-world robotics, where agents must interpret and reason in complex visual environments.

\textbf{Limitation} We only release YouTube video identifiers rather than full video files to respect licensing and platform terms. We have archived local copies, but sharing them for reproducibility depends on creator permission and compliance with applicable licenses.

\section*{Acknowledgement}
Shuo Wen, Edwin Meriaux and Chloe Si are supported by the FRQNT scholarship (The Fonds de recherche du Québec – Nature et technologies). The research is supported by the Natural Sciences and Engineering Research Council of Canada (NSERC) and computing infrastructure at MILA - Quebec AI Institute. The authors are grateful to the anonymous reviewers and Area Chairs for their thoughtful comments during the review of the paper.

\bibliographystyle{plainnat}
\bibliography{references}

@inproceedings{wang2025lvbench,
  title={Lvbench: An extreme long video understanding benchmark},
  author={Wang, Weihan and He, Zehai and Hong, Wenyi and Cheng, Yean and Zhang, Xiaohan and Qi, Ji and Ding, Ming and Gu, Xiaotao and Huang, Shiyu and Xu, Bin and others},
  booktitle={Proceedings of the IEEE/CVF International Conference on Computer Vision},
  pages={22958--22967},
  year={2025}
}

@inproceedings{xie2024funqa,
  title={Funqa: Towards surprising video comprehension},
  author={Xie, Binzhu and Zhang, Sicheng and Zhou, Zitang and Li, Bo and Zhang, Yuanhan and Hessel, Jack and Yang, Jingkang and Liu, Ziwei},
  booktitle={European Conference on Computer Vision},
  pages={39--57},
  year={2024},
  organization={Springer}
}

@inproceedings{lin2024video,
  title={Video-llava: Learning united visual representation by alignment before projection},
  author={Lin, Bin and Ye, Yang and Zhu, Bin and Cui, Jiaxi and Ning, Munan and Jin, Peng and Yuan, Li},
  booktitle={Proceedings of the 2024 conference on empirical methods in natural language processing},
  pages={5971--5984},
  year={2024}
}

@article{cheng2024videollama,
  title={Videollama 2: Advancing spatial-temporal modeling and audio understanding in video-llms},
  author={Cheng, Zesen and Leng, Sicong and Zhang, Hang and Xin, Yifei and Li, Xin and Chen, Guanzheng and Zhu, Yongxin and Zhang, Wenqi and Luo, Ziyang and Zhao, Deli and others},
  journal={arXiv preprint arXiv:2406.07476},
  year={2024}
}

@article{zhang2024llava,
  title={Llava-video: Video instruction tuning with synthetic data},
  author={Zhang, Yuanhan and Wu, Jinming and Li, Wei and Li, Bo and Ma, Zejun and Liu, Ziwei and Li, Chunyuan},
  journal={arXiv preprint arXiv:2410.02713},
  year={2024}
}

@article{chen2024longvila,
  title={Longvila: Scaling long-context visual language models for long videos},
  author={Chen, Yukang and Xue, Fuzhao and Li, Dacheng and Hu, Qinghao and Zhu, Ligeng and Li, Xiuyu and Fang, Yunhao and Tang, Haotian and Yang, Shang and Liu, Zhijian and others},
  journal={arXiv preprint arXiv:2408.10188},
  year={2024}
}

@article{bai2025qwen3,
  title={Qwen3-vl technical report},
  author={Bai, Shuai and Cai, Yuxuan and Chen, Ruizhe and Chen, Keqin and Chen, Xionghui and Cheng, Zesen and Deng, Lianghao and Ding, Wei and Gao, Chang and Ge, Chunjiang and others},
  journal={arXiv preprint arXiv:2511.21631},
  year={2025}
}

@inproceedings{li2024mvbench,
  title={Mvbench: A comprehensive multi-modal video understanding benchmark},
  author={Li, Kunchang and Wang, Yali and He, Yinan and Li, Yizhuo and Wang, Yi and Liu, Yi and Wang, Zun and Xu, Jilan and Chen, Guo and Luo, Ping and others},
  booktitle={Proceedings of the IEEE/CVF Conference on Computer Vision and Pattern Recognition},
  pages={22195--22206},
  year={2024}
}

@inproceedings{liu2024tempcompass,
  title={Tempcompass: Do video llms really understand videos?},
  author={Liu, Yuanxin and Li, Shicheng and Liu, Yi and Wang, Yuxiang and Ren, Shuhuai and Li, Lei and Chen, Sishuo and Sun, Xu and Hou, Lu},
  booktitle={Findings of the Association for Computational Linguistics: ACL 2024},
  pages={8731--8772},
  year={2024}
}

@inproceedings{fu2025video,
  title={Video-mme: The first-ever comprehensive evaluation benchmark of multi-modal llms in video analysis},
  author={Fu, Chaoyou and Dai, Yuhan and Luo, Yongdong and Li, Lei and Ren, Shuhuai and Zhang, Renrui and Wang, Zihan and Zhou, Chenyu and Shen, Yunhang and Zhang, Mengdan and others},
  booktitle={Proceedings of the IEEE/CVF conference on computer vision and pattern recognition},
  pages={24108--24118},
  year={2025}
}

@inproceedings{zhou2025mlvu,
  title={Mlvu: Benchmarking multi-task long video understanding},
  author={Zhou, Junjie and Shu, Yan and Zhao, Bo and Wu, Boya and Liang, Zhengyang and Xiao, Shitao and Qin, Minghao and Yang, Xi and Xiong, Yongping and Zhang, Bo and others},
  booktitle={Proceedings of the IEEE/CVF Conference on Computer Vision and Pattern Recognition},
  pages={13691--13701},
  year={2025}
}

@article{wu2024longvideobench,
  title={Longvideobench: A benchmark for long-context interleaved video-language understanding},
  author={Wu, Haoning and Li, Dongxu and Chen, Bei and Li, Junnan},
  journal={Advances in Neural Information Processing Systems},
  volume={37},
  pages={28828--28857},
  year={2024}
}

@article{chen2024rextime,
  title={Rextime: A benchmark suite for reasoning-across-time in videos},
  author={Chen, Jr-Jen and Liao, Yu-Chien and Lin, Hsi-Che and Yu, Yu-Chu and Chen, Yen-Chun and Wang, Yu-Chiang F},
  journal={Advances in Neural Information Processing Systems},
  volume={37},
  pages={28662--28673},
  year={2024}
}

@article{chen2024cg,
  title={Cg-bench: Clue-grounded question answering benchmark for long video understanding},
  author={Chen, Guo and Liu, Yicheng and Huang, Yifei and He, Yuping and Pei, Baoqi and Xu, Jilan and Wang, Yali and Lu, Tong and Wang, Limin},
  journal={arXiv preprint arXiv:2412.12075},
  year={2024}
}

@article{cheng2025video,
  title={Video-Holmes: Can MLLM Think Like Holmes for Complex Video Reasoning?},
  author={Cheng, Junhao and Ge, Yuying and Wang, Teng and Ge, Yixiao and Liao, Jing and Shan, Ying},
  journal={arXiv preprint arXiv:2505.21374},
  year={2025}
}

@article{zhu2025mmr,
  title={MMR-V: What's Left Unsaid? A Benchmark for Multimodal Deep Reasoning in Videos},
  author={Zhu, Kejian and Jin, Zhuoran and Yuan, Hongbang and Li, Jiachun and Tu, Shangqing and Cao, Pengfei and Chen, Yubo and Liu, Kang and Zhao, Jun},
  journal={arXiv preprint arXiv:2506.04141},
  year={2025}
}

@inproceedings{yu2025vrbench,
  title={Vrbench: A benchmark for multi-step reasoning in long narrative videos},
  author={Yu, Jiashuo and Wu, Yue and Chu, Meng and Ren, Zhifei and Huang, Zizheng and Chu, Pei and Zhang, Ruijie and He, Yinan and Li, Qirui and Li, Songze and others},
  booktitle={Proceedings of the IEEE/CVF International Conference on Computer Vision},
  pages={21655--21666},
  year={2025}
}

@article{fu2026video,
  title={Video-MME-v2: Towards the Next Stage in Benchmarks for Comprehensive Video Understanding},
  author={Fu, Chaoyou and Yuan, Haozhi and Dong, Yuhao and Zhang, Yi-Fan and Shen, Yunhang and Hu, Xiaoxing and Li, Xueying and Su, Jinsen and Long, Chengwu and Xie, Xiaoyao and others},
  journal={arXiv preprint arXiv:2604.05015},
  year={2026}
}

@inproceedings{li2022representation,
  title={From representation to reasoning: Towards both evidence and commonsense reasoning for video question-answering},
  author={Li, Jiangtong and Niu, Li and Zhang, Liqing},
  booktitle={Proceedings of the IEEE/CVF conference on computer vision and pattern recognition},
  pages={21273--21282},
  year={2022}
}

@article{zhang2025videollama,
  title={Videollama 3: Frontier multimodal foundation models for image and video understanding},
  author={Zhang, Boqiang and Li, Kehan and Cheng, Zesen and Hu, Zhiqiang and Yuan, Yuqian and Chen, Guanzheng and Leng, Sicong and Jiang, Yuming and Zhang, Hang and Li, Xin and others},
  journal={arXiv preprint arXiv:2501.13106},
  year={2025}
}

@article{bai2025qwen25vl,
  title={Qwen2.5-VL Technical Report},
  author={Bai, Shuai and Chen, Keqin and Liu, Xuejing and Wang, Jialin and Ge, Wenbin and Song, Sibo and Dang, Kai and Wang, Peng and Wang, Shijie and Tang, Jun and others},
  journal={arXiv preprint arXiv:2502.13923},
  year={2025}
}

@article{zhu2025internvl3,
  title={Internvl3: Exploring advanced training and test-time recipes for open-source multimodal models},
  author={Zhu, Jinguo and Wang, Weiyun and Chen, Zhe and Liu, Zhaoyang and Ye, Shenglong and Gu, Lixin and Tian, Hao and Duan, Yuchen and Su, Weijie and Shao, Jie and others},
  journal={arXiv preprint arXiv:2504.10479},
  year={2025}
}

@article{hong2025glm,
  title={Glm-4.5 v and glm-4.1 v-thinking: Towards versatile multimodal reasoning with scalable reinforcement learning},
  author={Hong, Wenyi and Yu, Wenmeng and Gu, Xiaotao and Wang, Guo and Gan, Guobing and Tang, Haomiao and Cheng, Jiale and Qi, Ji and Ji, Junhui and Pan, Lihang and others},
  journal={arXiv preprint arXiv:2507.01006},
  year={2025}
}

@article{wang2025internvl3,
  title={Internvl3. 5: Advancing open-source multimodal models in versatility, reasoning, and efficiency},
  author={Wang, Weiyun and Gao, Zhangwei and Gu, Lixin and Pu, Hengjun and Cui, Long and Wei, Xingguang and Liu, Zhaoyang and Jing, Linglin and Ye, Shenglong and Shao, Jie and others},
  journal={arXiv preprint arXiv:2508.18265},
  year={2025}
}

@article{yu2025minicpm,
  title={Minicpm-v 4.5: Cooking efficient mllms via architecture, data, and training recipe},
  author={Yu, Tianyu and Wang, Zefan and Wang, Chongyi and Huang, Fuwei and Ma, Wenshuo and He, Zhihui and Cai, Tianchi and Chen, Weize and Huang, Yuxiang and Zhao, Yuanqian and others},
  journal={arXiv preprint arXiv:2509.18154},
  year={2025}
}

@article{wang2025internvideo2,
  title={Internvideo2. 5: Empowering video mllms with long and rich context modeling},
  author={Wang, Yi and Li, Xinhao and Yan, Ziang and He, Yinan and Yu, Jiashuo and Zeng, Xiangyu and Wang, Chenting and Ma, Changlian and Huang, Haian and Gao, Jianfei and others},
  journal={arXiv preprint arXiv:2501.12386},
  year={2025}
}

@article{shen2024longvu,
  title={Longvu: Spatiotemporal adaptive compression for long video-language understanding},
  author={Shen, Xiaoqian and Xiong, Yunyang and Zhao, Changsheng and Wu, Lemeng and Chen, Jun and Zhu, Chenchen and Liu, Zechun and Xiao, Fanyi and Varadarajan, Balakrishnan and Bordes, Florian and others},
  journal={arXiv preprint arXiv:2410.17434},
  year={2024}
}

@article{chen2025scaling,
  title={Scaling rl to long videos},
  author={Chen, Yukang and Huang, Wei and Shi, Baifeng and Hu, Qinghao and Ye, Hanrong and Zhu, Ligeng and Liu, Zhijian and Molchanov, Pavlo and Kautz, Jan and Qi, Xiaojuan and others},
  journal={arXiv preprint arXiv:2507.07966},
  year={2025}
}

@misc{mossvl,
  title         = {MOSS-VL Technical Report},
  author        = {Wang, Pengyu and Tan, Chenkun and Zhou, Shaojun and Zhou, Qirui and Chen, Yanxin and He, Xingyang and Zeng, Huazheng and Cheng, Jijun and Wang, Chenghao and Qian, Xiaomeng and Wang, Pengfei and Huang, Zhan and Gao, Shanqing and Huang, Wei and Cao, Longjun and Ran, Wu and Liu, Jie and Zhu, Changtai and Wang, Hongkai and Tian, Yixian and Liu, Chenghao and Ye, Zhen and Wang, Xinghao and Jiang, Botian and Feng, Guoguo and Fei, Zhaoye and Li, Ruixiao and Chen, Mingshu and Gao, Yang and Cheng, Qinyuan and Li, Shimin and Qiu, Xipeng},
  year          = {2026},
  eprint        = {2608.15045},
  archivePrefix = {arXiv},
  primaryClass  = {cs.CV},
  url           = {https://arxiv.org/abs/2608.15045}
}

@article{yang2025kwai,
  title={Kwai keye-vl 1.5 technical report},
  author={Yang, Biao and Wen, Bin and Ding, Boyang and Liu, Changyi and Chu, Chenglong and Song, Chengru and Rao, Chongling and Yi, Chuan and Li, Da and Zang, Dunju and others},
  journal={arXiv preprint arXiv:2509.01563},
  year={2025}
}

@article{yue2025mimo,
  title={Mimo-vl technical report},
  author={Yue, Zihao and Lin, Zhenru and Song, Yifan and Wang, Weikun and Ren, Shuhuai and Gu, Shuhao and Li, Shicheng and Li, Peidian and Zhao, Liang and Li, Lei and others},
  journal={arXiv preprint arXiv:2506.03569},
  year={2025}
}

@inproceedings{yang2026cambrian,
  title={Cambrian-s: Towards spatial supersensing in video},
  author={Yang, Shusheng and Yang, Jihan and Huang, Pinzhi and Brown, Ellis and Yang, Zihao and Yu, Yue and Tong, Shengbang and Zheng, Zihan and Xu, Yifan and Wang, Muhan and others},
  booktitle={International Conference on Learning Representations},
  volume={2026},
  pages={78185--78225},
  year={2026}
}

@misc{qwen36_35b_a3b,
    title = {{Qwen3.6-35B-A3B}: Agentic Coding Power, Now Open to All},
    url = {https://qwen.ai/blog?id=qwen3.6-35b-a3b},
    author = {{Qwen Team}},
    month = {April},
    year = {2026}
}

@article{singh2025openai,
  title={Openai gpt-5 system card},
  author={Singh, Aaditya and Fry, Adam and Perelman, Adam and Tart, Adam and Ganesh, Adi and El-Kishky, Ahmed and McLaughlin, Aidan and Low, Aiden and Ostrow, AJ and Ananthram, Akhila and others},
  journal={arXiv preprint arXiv:2601.03267},
  year={2025}
}

@misc{deepmind2025gemini31pro,
  author = {{Google DeepMind}},
  title = {Gemini 3.1 Pro},
  year = {2026},
  howpublished = {\url{https://deepmind.google/models/model-cards/gemini-3-1-pro/}}
}

@misc{googledeepmind2026gemini31flashlite,
  author       = {{Google DeepMind}},
  title        = {{Gemini 3.1 Flash-Lite Model Card}},
  year         = {2026},
  howpublished = {\url{https://deepmind.google/models/model-cards/gemini-3-1-flash-lite/}},
}

@misc{openai2025gpt52,
  author       = {{OpenAI}},
  title        = {{Introducing GPT-5.2}},
  year         = {2025},
  howpublished = {\url{https://openai.com/index/introducing-gpt-5-2/}},
}

@misc{qwenteam2026qwen36plus,
  author       = {{Qwen Team}},
  title        = {{Qwen3.6-Plus: Towards Real World Agents}},
  year         = {2026},
  howpublished = {\url{https://qwen.ai/blog?id=qwen3.6}},
}

@misc{moonshot2026kimik26,
  author       = {{Moonshot AI}},
  title        = {{Kimi K2.6}: Advancing Open-Source Coding},
  year         = {2026},
  howpublished = {\url{https://www.kimi.com/blog/kimi-k2-6}},
}

@article{ali2025mcqa,
  title={Benchmarking and Mitigating MCQA Selection Bias of Large Vision-Language Models},
  author={Ali, Md. Atabuzzaman and Asgarov, Ali and Thomas, Chris},
  journal={arXiv preprint arXiv:2509.16805},
  year={2025}
}

@inproceedings{loginova2025addressing,
  title={Addressing blind guessing: Calibration of selection bias in multiple-choice question answering by video language models},
  author={Loginova, Olga and Bezrukov, Oleksandr and Shekhar, Ravi and Kravets, Alexey},
  booktitle={Proceedings of the 63rd Annual Meeting of the Association for Computational Linguistics (Volume 1: Long Papers)},
  pages={3216--3246},
  year={2025}
}

@article{zhang2025eob,
  title={Mitigating Easy Option Bias in Multiple-Choice Question Answering},
  author={Zhang, Hao and Li, Chen and Fernando, Basura},
  journal={arXiv preprint arXiv:2508.13428},
  year={2025}
}

@inproceedings{zhou2025se,
  title={SE-Jury: An LLM-as-Ensemble-Judge Metric for Narrowing the Gap with Human Evaluation in SE},
  author={Zhou, Xin and Kim, Kisub and Zhang, Ting and Weyssow, Martin and Gomes, Lu{\'\i}s F and Yang, Guang and Liu, Kui and Xia, Xin and Lo, David},
  booktitle={2025 40th IEEE/ACM International Conference on Automated Software Engineering (ASE)},
  pages={2606--2618},
  year={2025},
  organization={IEEE}
}

@article{zheng2023judging,
  title={Judging llm-as-a-judge with mt-bench and chatbot arena},
  author={Zheng, Lianmin and Chiang, Wei-Lin and Sheng, Ying and Zhuang, Siyuan and Wu, Zhanghao and Zhuang, Yonghao and Lin, Zi and Li, Zhuohan and Li, Dacheng and Xing, Eric and others},
  journal={Advances in neural information processing systems},
  volume={36},
  pages={46595--46623},
  year={2023}
}

\newpage

\appendix

\section{Magic as a testbed for cognitive visual reasoning}
\label{app:why_magic}

Professional magic performances provide a structured testbed for cognitive visual reasoning because their reveals are designed around a discrepancy between an expected outcome and a visually observed one. Explaining why a reveal is surprising requires more than recognizing objects or describing isolated actions. A model must identify the relevant visual evidence, track how the depicted situation changes over time, use commonsense knowledge to infer what would occur, and explain how the observed outcome departs from that expectation.

\paragraph{Expectation Formation and Violation}
Magic performances establish expectations through visible arrangements and actions, such as placing an object inside a container or covering it with a solid barrier. Interpreting the subsequent reveal requires applying commonsense physical and logical expectations, including object permanence, containment, and continuity. The model must then compare the observed outcome with the expected one and identify the specific discrepancy. Paradoxical Reasoning questions in our dataset formalize this process by requiring an explanation that connects the established setup, the ordinarily expected outcome, and the observed departure.

\paragraph{Temporal State Tracking}
The evidence needed to form an expectation may occur well before the reveal. Intervening actions, occlusions, and misdirection can separate the relevant observations across time. A model must therefore preserve object identities and states, track visible changes, and retrieve the earlier evidence relevant to the final outcome. This temporal integration is necessary to determine not only what occurred, but why the reveal is significant in light of the preceding performance.

\paragraph{Evidence Focus and Visual Grounding}
Magic also requires distinguishing reasoning-relevant evidence from visually noticeable but irrelevant actions introduced through misdirection. Because our dataset excludes audio narration and text captions, models must ground their explanations in the visible setup, actions, and reveal. A successful answer must connect these observations into a coherent logic chain rather than merely describe the final frame or state generically that the outcome is surprising.

\section{Why moving away from MC}
\label{anti_MC}

A natural design choice for a video-understanding benchmark is multiple-choice question answering (MCQA), since it offers transparent, deterministic scoring, and trivial aggregation. We deliberately reject this format for \emph{The Prestige}, because a growing body of evidence shows that MCQA systematically inflates the measured performance of Vision-Language Models (VLMs) without reflecting genuine visual understanding. We summarize the failure modes that motivated our open QA design.

\paragraph{Token and positional selection bias.}
VLMs exhibit a well-documented tendency to favor particular options and particular positions in the option list, independent of the visual evidence~\cite{ali2025mcqa,loginova2025addressing}. Prior work, which evaluates Qwen2.5-VL across difficulty-stratified MCQA benchmarks, reports that the model shows a preference toward the token \textbf{A} on easy items and towards the option appearing first on hard questions~\cite{ali2025mcqa}.

\paragraph{Shortcutting and uninspectable reasoning.}
Beyond token bias, MCQA difficulty is largely determined by its distractors, which let models eliminate implausible options using world knowledge alone without ever understanding the video content~\cite{chen2024cg}. Zhang et al. identify a related shortcut across six VQA benchmarks (including NExT-QA, STAR, and Video-MME) and four VLM families: providing only the vision input and the options (withholding the question entirely) recovers a mean accuracy of $51.57\%$, just $9.54$ points below the full ``vision+question+options'' setting~\cite{zhang2025eob}. Even when MCQA scoring is accurate, the format suppresses the reasoning trace we wish to evaluate. A correct option \textbf{A} reveals nothing about whether the model tracked the deck across a misdirection or fabricated a causal story, and an incorrect \textbf{A} reveals nothing about which step of the chain failed.

\paragraph{Why open-ended answers.}
The same properties that make MCQA brittle make open QA evaluation a natural fit for our setting. By forcing the model to commit to a specific causal account in its own words, the open-ended format eliminates the token, positional, and elimination shortcuts described above and exposes fabrication directly. A model that has not tracked the hidden state of the trick will either decline or hallucinate, and either failure is visible to the grader. Crucially, in the context of our AQ (Section~\ref{sec:qa-generation}), a free-form answer can surface sycophantic agreement with a false statement, whereas an MCQ over an AQ.FS can provide multiple clues to the assumption we are testing. This consideration has driven recent long-video benchmarks such as CG-Bench~\cite{chen2024cg} and VRBench~\cite{yu2025vrbench} to shift partially or wholly toward open-ended evaluation, and we adopt the same format throughout the dataset.

\section{Dataset comments}
\label{sec:datasetcomments}

Our dataset is smaller than general-purpose video-QA benchmarks. However, our goal is to evaluate cognitive visual reasoning through open-ended questions, where we prioritize video quality, question depth, and annotation reliability over dataset size.  The concentration in CR and PR is intentional and reflects the benchmark’s central objective: magic performances are especially effective for evaluating long-horizon state tracking and cognitive reasoning about outcomes that violate established expectations. The distribution is therefore weighted toward the target capabilities rather than designed to uniformly cover all video-understanding skills. 

We evaluate all models using frames sampled at 1 FPS due to model and API constraints. We encourage future users to evaluate the videos at their native frame rates to preserve fine-grained temporal information and make fuller use of the dataset.

\section{Quality Standards}

Across this dataset, high-quality videos and question-answer pairs are fundamental. In this section, we outline our Video Quality Standards (VQS) for source video collection and Question-Answer Quality Standards (QAQS) for dataset construction.

\paragraph{Video Quality Standard: }
\label{sec:vqc}
\begin{itemize}
    \item \textbf{Magic quality} The video must depict a professional magic performance. Amateur or homemade recordings are excluded, as they tend to exhibit performance flaws, insufficient complexity, and unnatural pacing. 

    \item \textbf{Ethical quality} The selected video must not contain content that is violent, unethical, or otherwise inappropriate. This applies to both the magician and the platforms presenting the act.

    \item \textbf{Temporal length.} The magic performance must exceed $90$ seconds in duration, presented in its full, unclipped form (The whole video could be longer). This lower bound ensures the inclusion of multi-stage routines and long-range foreshadowing, thereby stress-testing the temporal reasoning, state tracking, and memory abilities of the models. Consequently, VLMs are required to autonomously identify the proper performance window and reason over it without external segmentation.

    \item \textbf{Visual fidelity.} Each video must be recorded at a minimum resolution of $720$p (most are $1080$p), under adequate lighting conditions. Human annotators verify that every visual cue required to answer the associated questions is clearly discernible. Videos exhibiting abrupt cuts, pronounced camera jitter, or rapid viewpoint switching are excluded.

    \item \textbf{Visual self-sufficiency.} The performance must be fully comprehensible from visual information alone, without reliance on audio narration or dialogue. Videos containing hardcoded captions are excluded, ensuring that evaluation reflects genuine visual understanding rather than incidental textual cues.

    \item \textbf{Minimal text:} All videos contain minimal on-screen text and no captions to prevent models from over-relying on textual cues instead of visual understanding.

    \item \textbf{Causal coherence.} The video must exhibit a clear causal structure and present commonsense-violating surprises that are intentionally orchestrated by the performer. Surprises arising from technical glitches, editing artifacts, or other factors outside the performer's control are excluded.

    \item \textbf{Contextual density.} Each video must be visually rich and contain multiple distinct magic surprises, ensuring that the associated question--answer pairs remain non-trivial and that the benchmark probes a model's capacity to process dense, long-horizon visual evidence.

\end{itemize}

\paragraph{Question-Answer Quality Standard:}
\label{QAQS}
\begin{itemize}
    \item \textbf{Triviality filter.} Questions answerable from the question text alone, without watching the video, 
    are removed.
    \item \textbf{Insufficient-context filter.} Questions whose visual grounding is impossible (relevant frames off-screen, 
    motion-blurred beyond recognition, occluded by the performer's body, or cut away by the camera) are removed. A careful 
    human viewer with access to the cited timestamp must be able to answer from the visual content alone.

    \item \textbf{External-knowledge filter.} Questions requiring outside knowledge (a magician's biography, an unobservable
     backstage detail, magic-industry trivia) are removed, so that performance reflects video understanding rather than 
     knowledge retrieval.

    \item \textbf{Grammar and phrasing.} Each question is checked for spelling, subject and verb agreement, punctuation, 
    and natural English phrasing.

    \item \textbf{Timestamp alignment.} The frames at and before the cited timestamp must contain the visual evidence 
    required to answer the question. Items whose answer is only visible after the cited timestamps are corrected to the 
    appropriate frame range.

    \item \textbf{Answer correctness.} The reviewer answers the question independently from the video before consulting 
    the proposed answer, and compares the two. Disagreements are resolved by re-watching the window together; persistent 
    disagreements escalate to a third adjudicator.

    \item \textbf{Question-type matching.} The reviewer confirms that the question genuinely requires the reasoning 
    targeted by its declared type (\textit{contextual reasoning} for multi-step inference over a window, 
    \textit{instant reasoning} for single-frame recognition), and reclassifies items that are mismatched to their type.

    \item \textbf{Stylistic consistency.} Answers are checked against an in-house style guide: declarative sentences, 
    present tense, third-person reference to the performer, no hedging language; numeric answers use digits, named props 
    use the spelling agreed upon in the keyword vocabulary, and card suits and ranks use a fixed canonical form.

\end{itemize}

\section{Dataset Generation Pipeline}

\label{data_pipeline}

This section in the appendix details the end-to-end pipeline by which the dataset was constructed, 
from raw video sourcing to the final question--answer records. We utilize the VQS and QAQS outlined in Appendix~\ref{sec:vqc} for dataset generation to maintain consistent high quality throughout the dataset. This process begins with sourcing the videos in Appendix~\ref{sec:sourcing}, and then walks through the 
two-pass annotation pipeline used to produce QA pairs (Appendix~\ref{AP}). Then goes to peer 
review that edits and validates QA pairs to make sure they meet the QAQS (Appendix~\ref{Peer_review}), and the final schema 
under which every item is stored (Appendix~\ref{scheme}). Together, these stages are designed to ensure that each 
benchmark item is visually grounded, unambiguously answerable, and aligned with the reasoning skill it is intended to probe.

\subsection{Video Sourcing Protocol}

\label{sec:sourcing}

We construct the dataset from publicly available online professional magic performances. An 
initial sweep yielded a pool of several hundred candidates covering a large diversity of magicians and acts.

Each video was then manually inspected end-to-end against the VQS defined in Appendix~\ref{sec:vqc},
with no automated pre-filtering applied. A trained annotator watched the full clip and recorded a per-criterion verdict on
the VQS dimensions. A candidate was retained only if it satisfied every criterion; failure on 
any single dimension was sufficient grounds for exclusion. The most common rejection causes were reliance on verbal or audio-cued reveals (failing visual self-sufficiency), aggressive editing with cuts or slow-motion replays (failing visual fidelity), and short single-effect clips lacking multiple distinct surprises (failing contextual density). Borderline cases were escalated to a second reviewer. All reviews were conducted by watching the video clips without any audio to faithfully verify if the underlying acts were understandable without audio.

\subsection{Question Generation Protocol}
\label{AP}
\paragraph{First pass: video scan.} An annotator conducts an audio-free scan through the video to fully watch the events
and get a contextual understanding of the video. The annotator does not stop the video during this phase and
simply writes down a log of frames where the questions can be generated. The frames are at times in the video
when non-trivial changes of object state occur 
(an object disappearing, materializing, transforming, or relocating). This yields a dense list of candidate moments 
at frame-level resolution, capturing the trick's branching structure exhaustively before any question is committed 
to writing. The annotator is allowed to restart the video once it has ended if any important event was missed in the
log.

\paragraph{Second pass: question authoring.} The annotator at this point has seen the full video and has built a full
log of question-worthy time points in the video. The annotator then starts to draft question-and-answer pairs, choosing the question type that best probes the reasoning demanded by that moment 
(see Section~\ref{sec:qa-generation}). Each question is anchored by an explicit \textit{timestamp} 
(e.g., ``At 0:57 of the video, $\dots$'') so that the model's context window is unambiguously defined at evaluation 
time. The annotator additionally tags the item with one or more \textit{keywords} naming the dominant prop or 
technique (e.g., \texttt{Card Trick}, \texttt{Coin Trick}, \texttt{Illusions}, \texttt{Flying Trick}, 
\texttt{Rubik's Cube Trick}; sample vocabulary in Figure~\ref{fig:word_cloud}).

\begin{figure}[ht]
    \centering
    \includegraphics[width=1\linewidth]{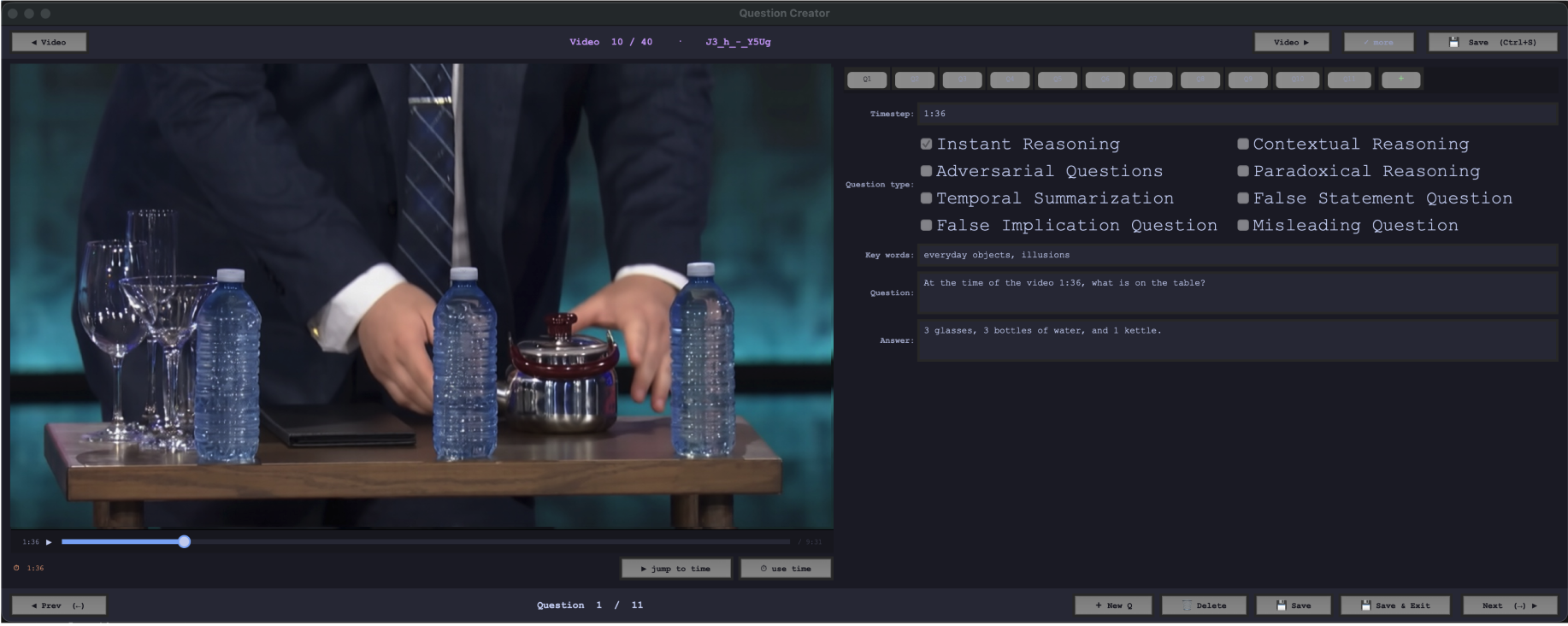}
    \caption{Sample question generation with generation tool.}
    \label{Q_gen}
\end{figure}

A visual example of the custom tool used to generate the questions can be seen in Figure~\ref{Q_gen}.

\subsection{Peer Review Protocol}
\label{Peer_review}

At this stage, a group of video collectors and annotators obtained magic videos, eliminated those that did not meet the VQS, and then generated a series of QA pairs. All questions for a given video were generated by the same annotator. To validate the quality of the data, a peer review is conducted to make
sure all questions meet the QAQS.

A record is admitted into the final benchmark only after passing every item on the checklist. Disagreements unresolved by the original annotator and the reviewer escalate to a third annotator whose decision is final and is logged in the dataset's audit trail.

\subsection{Final Record Schema}
\label{scheme}
As the QA pairs are finalized and meet the required standard, the output is structured as a single JSON file keyed by source video identifier. Each video maps to a \texttt{Questions} list of question--answer records together with a \texttt{metadata} block, so that all items grounded in the same performance are colocated and can be retrieved jointly at evaluation time. Formally, the dataset has the structure
\begin{center}
\texttt{\{video\_id: \{"Questions": [record$_1$, \dots, record$_n$], "metadata": \{\dots\}\}\}},
\end{center}
where each record is a tuple of the form
\begin{center}
\texttt{(UID, Timestamp, Question type, Key words, Question, Answer)}.
\end{center}

\paragraph{Field semantics.}
The \texttt{video\_id} corresponds to the YouTube identifier of the source clip and serves as the primary key for retrieving the underlying video. Each video carries a \texttt{metadata} block whose \texttt{start} field marks the timestamp at which the magic performance proper begins, allowing the evaluation protocol to skip pre-performance introductions and stage setup. However, this data is not provided to the model for question answering. The \texttt{UID} is a globally unique integer identifier for the item, assigned independently of the video grouping so that records can be referenced unambiguously across splits and analyses. The \texttt{Timestamp} anchors the question to a specific frame range in the source video and defines the model's context window at evaluation time. The \texttt{Question type} is drawn from the closed taxonomy introduced in Appendix~\ref{sec:qa-generation}. The \texttt{Key words} field is the keyword of the dominant prop or technique. Finally, \texttt{Question} and \texttt{Answer} are as defined in Section~\ref{AP}.

\paragraph{Example record.}
A representative entry, abbreviated for space, is shown below.
\begin{lstlisting}[language=json,basicstyle=\ttfamily\footnotesize,breaklines=true,columns=fullflexible]
{
  "RSeab6QqZok": {
        "Questions": [
            {
                "UID": 2412,
                "Timestep": "1:25",
                "Question type": "Paradoxical Reasoning",
                "Key words": "Body Magic Trick",
                "Question": "At 1:25 of the video, why are the people surprised?",
                "Answer": "The people are surprised because the three legs can move like they are real."
            },
      ...
            {
                "UID": 2418,
                "Timestep": "3:28",
                "Question type": "Paradoxical Reasoning",
                "Key words": "Body Magic Trick",
                "Question": "At 3:28 of the video, why are the people surprised?",
                "Answer": "The people are surprised because this man's legs are detached, but he's still moving back using those three legs."
            }
        ],
        "metadata": {
            "start": "0:00"
        }
    }
}
\end{lstlisting}

\section{Question Types}

\label{sec:question-types}

This appendix expands the question types introduced in Section~\ref{sec:qa-generation}. For each of the question types, we specify the target ability the question isolates, the failure modes it is designed to surface, and its position relative to adjacent categories of questions.

\subsection{Instant Reasoning (IR)}


Instant Reasoning questions probe the model's ability to extract a complete, query-relevant description of the visual state at a single anchored timestamp $t$. The reasoning window is deliberately collapsed to just one frame. Although the model receives the video context as input, the answer must be grounded in evidence visible at $t$.

The category serves a structural role in the ability of VLMs to comprehend videos. It isolates individual frames on which Contextual and Paradoxical Reasoning depend. If a model fails IR on a given frame, any downstream reasoning anchored to objects or events related to that frame is unreliable regardless of its long-context capabilities. This type of question also evaluates temporal slippage failure, where the model answers the question from a nearby but incorrect frame. This can lead to incorrect answers as those frames contain different information.

\subsection{Contextual Reasoning (CR)}

Contextual Reasoning (CR) questions ask the model to answer a query at a specific timestamp using evidence established earlier in the video. The queried frame is insufficient on its own. The answer may depend on the earlier identity of the object, a state tracked through occlusion, a prior view of a surface now hidden, or a persistent fact such as which prop was chosen or which card was signed. The model must therefore maintain a cross-frame memory of the performance and selectively retrieve the subset of that memory relevant to the query, fusing the aggregated context to produce a grounded answer.

CR questions cannot be resolved from the queried frame alone. The model must connect objects and their states across time and retrieve the earlier evidence to answer. This focus on reasoning about what is true distinguishes CR from Paradoxical Reasoning (PR), which additionally asks what should have happened and why the observed outcome was unexpected. CR also differs from Temporal Summarization (TS) by answering a specific query rather than summarizing events over a time window. Common failure modes include confusing similar props introduced at different times, relying on the most recent view instead of the relevant earlier evidence, and confidently answering while overlooking a fact established much earlier.

\subsection{Paradoxical Reasoning (PR)}

A PR question asks why an event at a particular timestamp is unexpected to a human audience. To answer the question, the model must use the preceding video to infer what should happen under ordinary physical and causal constraints, identify what actually happens, and explain the discrepancy. This distinguishes PR from CR, which can be answered by accurately tracking the observed state alone. Stating either the expected or the observed outcome in isolation does not explain the surprise.

Forming that expectation requires the model to track relevant objects through occlusion and misdirection and apply principles such as object permanence, gravity, identity under transformation, and spatiotemporal continuity. These steps can fail independently. A model may track the objects correctly but never form a prediction, leaving it with an accurate description rather than an explanation. Or it may base a plausible prediction on the wrong earlier state, producing an explanation that sounds coherent but is factually incorrect.

Magic poses a further challenge that a seemingly minor action may establish an expectation long before the reveal, with misdirection filling the interval. To explain the surprise, the model must trace the reveal back to that action and identify when and how the expectation arose. PR requires the model to pay attention to details and trace the backward causal links, testing a capability that single-frame and short-horizon evaluations cannot capture.

\textbf{Question framing:} We ask ``Why are the people watching surprised?'' rather than ``What is surprising?'' to make the audience's perspective explicit. A model trained with magic videos may recognize the method behind a trick and find the reveal entirely trivial and nothing surprising. Asking it to identify what is surprising without specifying whose expectations would therefore leave the task ambiguous. Our formulation keeps PR meaningful even for models that can decipher the illusion. To answer correctly, the model must reconstruct what the performance leads a human viewer to believe, what the viewer would expect on that basis, and why the reveal conflicts with that expectation. This is essentially the flow of cognition and requires reasoning from the audience's understanding of the preceding events, even when the model's own knowledge supports a different prediction. The ability to decipher a trick thus does not make PR trivial. The model needs to possess the cognitive process of human viewers and explain how the performance creates a contradiction for its audience.

\subsection{False Implication Question (AQ.FI)}

False Implication Questions (AQ.FI) test whether a model can identify and reject a false presupposition embedded in a question. Rather than explicitly making a false claim, these questions use their wording or structure to inject false information. A correct response must uncover the implicit assumption, check it against the video, and why it is false. AQ.FI reveals failures to recognize and reject embedded false claims and to ground reasoning in visual evidence.

AQ.FI and AQ.FS both test whether models verify premises of a question before reasoning from them. Their difference lies in how false information is introduced. AQ.FI embeds it implicitly in the question, whereas AQ.FS asserts it explicitly. We design these questions to test the robustness of the visual reasoning, revealing whether models can evaluate the premises of a question rather than simply answer without a second thought.

\subsection{False Statement Question (AQ.FS)}

False Statement Questions (AQ.FS) test whether a model can reject an explicitly asserted false premise with video evidence. Each question explicitly presents a false factual claim followed by a question that assumes the claim is true. A correct response should reject the presented premise and re-ground its reasoning on actual visual information. An incorrect response, on the other hand, often accepts the false premise and hallucinates details to make the answer appear to be coherent.

AQ.FS questions target a failure mode in which a model accepts a false claim despite contradictory video evidence. Multiple-choice evaluation could obscure this failure by allowing models to exploit reasoning shortcuts by providing diverse samples. The open-ended format, by contrast, exposes the entire causal chain of the responses of the model.

AQ.FS questions are not a simple correctness test. The premise in these questions is not arbitrarily wrong, but chosen to be verifiable from visual context. A successful answer therefore requires the model to (i) recognize the prompt as making a false factual claim about visual content, (ii) retrieve the relevant visual evidence, (iii) detect the conflict between the asserted premise and the retrieved evidence, and (iv) reject the premises and perform grounded reasoning.

\subsection{Temporal Summarization (TS)}

\label{sec_sub:ts}


Temporal Summarization questions require the model to produce a compact, abstract description of the events within a specified temporal window. A correct TS response captures essential details, while suppressing both low-level enumerations and over-generalizations. The category isolates the ability to compress a temporal interval rather than the ability to enumerate.

TS questions mainly probe two characteristic failure modes. The first is granularity misalignment, in which the model either provides a frame-by-frame narration or overgeneralizes (``the magician performs a card trick''). The second is event omission, in which the model produces a locally accurate summary but skips a critical action within the window, often the action that motivated the question. Both failure modes are difficult to assess by multiple-choice evaluation, where the candidate answers supply predefined descriptions rather than requiring the model to summarize critical events and choose its own level of abstraction.

\section{Grading Pipeline}
\label{sec:grading_pipeline}

This section describes the human and machine grading protocols used to evaluate VLM responses. Human grading provides the primary benchmark scores, while machine grading provides a separate evaluation for comparison with human judgments.

\subsection{Grading Protocol}

After collecting the model responses, we evaluated them using both human grading and a majority vote among three API-based VLM judges. We report the two sets of scores separately and compare them across all 14 models and six question categories.

\begin{figure}[ht]
    \centering
    \includegraphics[width=1\linewidth]{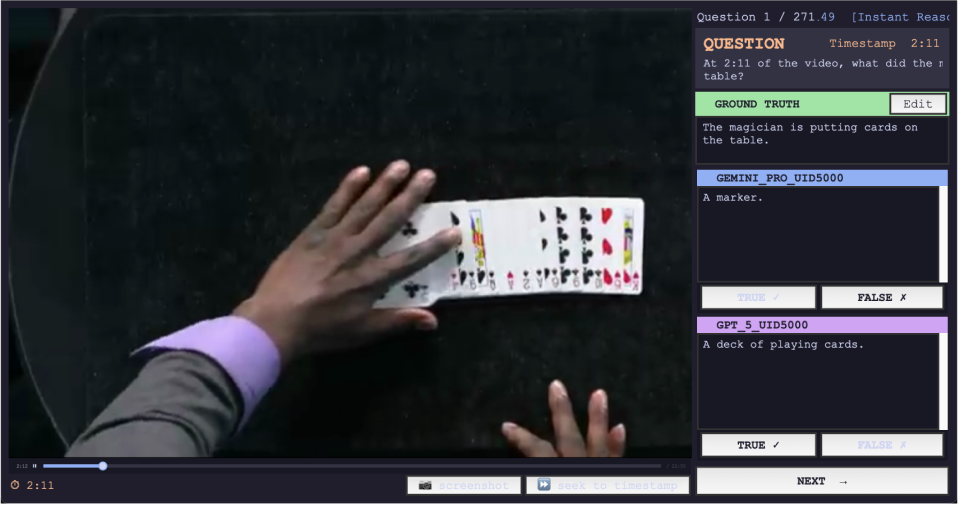}
    \caption{Custom interface used by human evaluators to grade VLM responses.}
    \label{grading_tool}
\end{figure}

The two grading protocols are described below:
\begin{itemize}
    \item \textbf{Automated Machine Grading:} We use an LLM-as-a-judge approach~\cite{zheng2023judging,zhou2025se} with three graders: \texttt{qwen3.7-plus-2026-05-26}, \texttt{gemini-3.5-flash-lite}, and \texttt{gpt-5.6-luna}. All graders are accessed through their official APIs with temperature 0.0 and seed 42. Each grader independently evaluates every model response using the corresponding video frames, question, ground-truth human answer, and question type, following the rubric in Figure~\ref{fig:grading-prompt}. Each grader assigns a binary score of 0 (Fail) or 1 (Pass) to the model answer. The final machine grade is determined by majority voting. A response passes if at least two of the three graders assign a score of 1. Complete machine-grading results are reported in Appendix~\ref{sec:machine_grading}.

    \item \textbf{Human Grading:} Human evaluators manually grade responses from all 14 models to obtain the primary benchmark scores. To support consistent grading across models, the workload was organized by question subsets rather than by model. Each evaluator was assigned a set of questions and graded every response of all the models to that same set. The custom grading interface used by human evaluators is shown in Figure~\ref{grading_tool}.
\end{itemize}

\begin{promptbox}
\footnotesize

You are a strict grading assistant. Your sole task is to compare each submitted answer against its ground-truth answer and assign a binary score.

\medskip
\textbf{Inputs}

For each grading request, you will receive:
\begin{enumerate}
    \item Video frames.
    \item A structured data object containing the questions, ground-truth answers, submitted answers, and question types associated with the video.
\end{enumerate}

The input follows this conceptual structure:

\begin{lstlisting}[style=prompt]
{
    video_id: {
        uid: {
            Question: question,
            GT_Answer: ground_truth_answer,
            Submitted_Answer: submitted_answer,
            Question_type: question_type
        }
    }
}
\end{lstlisting}

The top-level key is the video ID and identifies the corresponding provided video.

Each inner key is a question UID. Preserve all video IDs and UIDs exactly in your output.

Grade every UID independently.

Inspect the provided video whenever necessary to understand the question, identify visual events, or resolve ambiguous references such as pronouns, object identities, actions, timing, and spatial relationships.

\medskip
\textbf{Rule 1: Ground Truth Is the Absolute Authority}

The \texttt{GT\_Answer} must always be treated as completely correct.

Do not override, dispute, reinterpret, or correct the ground-truth answer based on your own interpretation of the question or video.

Use the video only to understand the context and determine whether the submitted answer matches the ground truth.

If your interpretation of the video conflicts with the ground-truth answer, always follow the ground-truth answer.

\medskip
\textbf{Rule 2: Binary Scoring}

Assign one of the following scores:
\begin{itemize}
    \item 1 means Pass. The submitted answer captures all key facts required by the ground-truth answer at the same level of specificity. Equivalent wording is acceptable.
    \item 0 means Fail. The submitted answer omits a required key fact, is insufficiently specific, contains an incorrect claim, or otherwise fails to fully satisfy the ground-truth answer.
\end{itemize}

An answer that is generally correct but omits a specific detail required by the ground truth must receive 0.

\medskip
\textbf{Rule 3: Strict Specificity}

The submitted answer must match the level of specificity of the ground-truth answer.

A broadly true statement is not sufficient when the ground truth provides a more specific answer.

\medskip
Example 1:
\begin{itemize}
    \item Question: Which hand is the ball in?
    \item Ground truth: The ball is in the man's right hand.
    \item Submitted answer: The ball is in his hand.
    \item Score: 0
    \item Reason: The submitted answer does not identify the right hand.
\end{itemize}

Example 2:
\begin{itemize}
    \item Question: How many Kings are on the table?
    \item Ground truth: There are four Kings on the table.
    \item Submitted answer: There are some cards on the table.
    \item Score: 0
    \item Reason: The submitted answer omits both the number and the identity of the cards.
\end{itemize}

Example 3:
\begin{itemize}
    \item Question: Why are the people surprised?
    \item Ground truth: The tricks were predicted and written on the board before the act started.
    \item Submitted answer: The magician performed a magic trick with a board, which surprised the people.
    \item Score: 0
    \item Reason: The submitted answer does not explain that the tricks were predicted and written down before the performance.
\end{itemize}

\medskip
\textbf{Rule 4: Question-Type Requirements}

Apply the rule corresponding exactly to the value of \texttt{Question\_type}.

\medskip
\textbf{Adversarial, False Implication}

Assign 1 only when the submitted answer provides both:
\begin{enumerate}
    \item The correct yes-or-no judgment.
    \item A sufficiently specific and correct explanation.
\end{enumerate}

If either component is missing, vague, or incorrect, assign 0.

\medskip
\textbf{Adversarial, False Statement}

Assign 1 only when the submitted answer both:
\begin{enumerate}
    \item Clearly rejects or corrects the false premise.
    \item Provides the correct replacement fact.
\end{enumerate}

If either component is missing, vague, or incorrect, assign 0.

\medskip
\textbf{Temporal Summarization}

Assign 1 only when the submitted answer gives a correct and complete summary of the period specified in the question.

Assign 0 when the summary is incomplete, vague, temporally incorrect, or omits a required event.

\medskip
\textbf{Paradoxical Reasoning}

Assign 1 only when the submitted answer gives specific and correct details explaining what is surprising, paradoxical, or unexpected.

A generic statement that something was surprising or magical is insufficient and must receive 0.

\medskip
\textbf{Instant Reasoning}

Assign 1 only when the submitted answer correctly identifies the content, object, person, action, state, or event described by the ground truth.

A wrong identification or materially different meaning must receive 0.

\medskip
\textbf{Contextual Reasoning}

Assign 1 only when the submitted answer correctly identifies the content, object, person, action, state, relationship, or event described by the ground truth using the relevant video context.

A wrong identification or materially different meaning must receive 0.

\medskip
\textbf{Rule 5: Additional Grading Rules}

\begin{itemize}
    \item Do not penalize differences in wording, grammar, sentence structure, or vocabulary when the meaning is equivalent.
    \item Do not require exact string matching.
    \item Do not penalize the omission of minor, nonessential descriptors, such as omitting wooden from wooden table, when the core object and action remain correct.
    \item Do not accept vague answers that avoid making an incorrect claim but fail to answer the question specifically.
    \item Assign 0 if the submitted answer includes clearly false, contradictory, or hallucinated information, even when the rest of the answer is correct.
    \item Assign 0 if an added claim materially changes or contradicts the correct answer.
    \item Do not award partial credit.
    \item Do not use information from one UID to grade another UID.
    \item Judge only whether the submitted answer satisfies the question and ground-truth answer for that UID.
\end{itemize}

\medskip
\textbf{Output Requirements}

Return exactly one syntactically valid JSON object.

Do not include Markdown code fences, introductory text, concluding text, analysis, or comments outside the JSON object.

Provide one entry for every input video ID and every UID.

For each UID:
\begin{itemize}
    \item Include a field named \texttt{reason} containing exactly one concise sentence explaining why the answer passed or failed.
    \item Include a field named \texttt{score} containing the integer 0 or 1.
    \item The reason must identify the decisive matched, missing, incorrect, vague, or hallucinated detail.
    \item Do not provide lengthy reasoning or hidden chain-of-thought.
\end{itemize}

The output must follow this conceptual structure:

\begin{lstlisting}[style=prompt]
{
    video_id: {
        uid: {
            reason: one concise sentence explaining the decision,
            score: 0 or 1
        }
    }
}
\end{lstlisting}

The actual response must use standard valid JSON syntax, including the quotation marks required by JSON.

Additional output constraints:
\begin{itemize}
    \item Do not include trailing commas.
    \item Do not omit any video ID or UID.
    \item Do not add video IDs or UIDs that were not present in the input.
    \item Preserve the original video IDs and UIDs exactly.
    \item The score must be an integer rather than a string.
\end{itemize}

The video frames and corresponding grading data will be provided immediately after this instruction.

\end{promptbox}
\captionof{figure}{Prompt used by our LLM-based grading assistant. The grader receives video frames and the corresponding questions, ground-truth answers, submitted answers, and question types. It independently assigns each answer a binary score and returns a JSON object with one concise reason and score per question UID.}
\label{fig:grading-prompt}

\section{Evaluation Prompt}

We provide the evaluation prompt used in our experiments below. The \texttt{<think>} component serves only as a diagnostic record of the reasoning of the model and is excluded from grading. Only the response extracted from the \texttt{<answer>} tags is graded. We intentionally restrict the model perspective to prevent it from deciphering the magic. We also added extra guidance due to the limitations of the open-source models that we test. However, we encourage future users of our dataset to explore alternative prompts to investigate how task framing affects model performance.

\begin{tcolorbox}[
  enhanced,
  colback=black!2,
  colframe=black!62,
  colbacktitle=black!62,
  coltitle=white,
  title={Model Evaluation Prompt},
  fonttitle=\normalsize\bfseries,
  fontupper=\small,
  boxrule=0.8pt,
  arc=3mm,
  left=10pt, right=10pt,
  top=8pt, bottom=8pt,
  toptitle=5pt, bottomtitle=5pt
]
\raggedright
\setlength{\parskip}{6pt}

You are a world expert in understanding videos and answering questions based on visual information. You are also an audience watching magic performances, who does NOT have any prior knowledge of any magic tricks. Your response must be clear, concise, and non-repetitive. Do not copy the template text. Do not repeat yourself. Do not try to decipher the magic to show your understanding. Only reason based on visual information.

Output exactly in this form:\\
\texttt{<think>}\\
Brief reasoning in plain sentences only.\\
\texttt{</think>}\\
\texttt{<answer>}\\
Answer\\
\texttt{</answer>}

Do not write anything outside these tags.\\
Stop immediately after \texttt{</answer>}.
\end{tcolorbox}

\section{Machine Grading Results}
\label{sec:machine_grading}

Category and overall accuracy are the percentages of responses that pass within each category and across the full dataset, respectively. Human grading remains the primary benchmark evaluation; Table~\ref{tab:human_machine_difference} reports the differences between human and majority-vote machine scores for all 14 models.

\begin{table*}[hbp]
    \centering
    \resizebox{\textwidth}{!}{
        \begin{tabular}{p{4.7cm} | c c c c c c c}
        \toprule
        \textbf{Model} \hfill \textbf{Size} & \textbf{IR} & \textbf{CR} & \textbf{PR} & \textbf{AQ.FI} & \textbf{AQ.FS} & \textbf{TS} & \textbf{Overall} \\
        \midrule

        \textit{\qquad Open-source models} & & & & & & & \\
        \midrule
        Qwen3VL-Thinking \citep{bai2025qwen3} \hfill 8B
        & 32.20 & 30.88 & 3.34 & 20.60 & 4.35 & 9.82 & 18.24 \\

        Qwen3VL-Instruct \citep{bai2025qwen3} \hfill 8B
        & 37.56 & 34.43 & 8.29 & 14.59 & 10.87 & 16.07 & 21.68 \\

        InternVideo2.5-Chat \citep{wang2025internvideo2} \hfill 8B
        & 23.41 & 23.25 & 1.20 & 3.00 & 1.09 & 1.79 & 11.48 \\

        InternVL3.5 \citep{wang2025internvl3} \hfill 8B
        & 17.56 & 23.63 & 1.20 & 15.45 & 6.52 & 1.79 & 12.63 \\

        MiMo-VL-SFT-2508 \citep{yue2025mimo} \hfill 7B
        & 14.15 & 19.31 & 0.80 & 14.59 & 6.52 & 4.46 & 10.66 \\

        MOSS-VL-Instruct \citep{mossvl} \hfill 11B
        & 32.20 & 30.50 & 7.22 & 24.89 & 1.09 & 11.61 & 19.84 \\

        Qwen3.6 \citep{qwen36_35b_a3b} \hfill 35A3B
        & 44.88 & 43.20 & 19.12 & 34.33 & 15.22 & 25.89 & 32.06 \\

        Cambrian-S \citep{yang2026cambrian} \hfill 7B
        & 18.05 & 19.19 & 1.34 & 12.02 & 0.00 & 4.46 & 10.61 \\

        Keye-VL-1.5 \citep{yang2025kwai} \hfill 8B
        & 24.88 & 24.90 & 3.34 & 20.60 & 7.61 & 8.04 & 15.43 \\

        Kimi-K2.6 \citep{moonshot2026kimik26} \hfill 1.1TA32B
        & 52.20 & 45.24 & 24.33 & 33.91 & 26.09 & 27.68 & 35.78 \\

        \midrule

        \textit{\qquad Proprietary models} & & & & & & & \\
        \midrule

        GPT-5.2 \citep{openai2025gpt52} \hfill UKN
        & 47.80 & 42.95 & 27.67 & 15.02 & 8.70 & 39.29 & 33.53 \\

        Gemini3.1-Flash-Lite \citep{googledeepmind2026gemini31flashlite} \hfill UKN
        & 47.32 & 46.25 & 29.28 & 37.77 & 17.39 & 35.71 & 37.85 \\

        Qwen3.6-Plus \citep{qwenteam2026qwen36plus} \hfill UKN
        & 43.90 & 48.03 & 26.07 & 29.18 & 14.13 & 34.82 & 35.97 \\

        Gemini3.1-Pro \citep{deepmind2025gemini31pro} \hfill UKN
        & 50.24 & 48.41 & 37.70 & 41.20 & 56.52 & 34.82 & 43.78 \\

        \bottomrule
        \end{tabular}
    }

    \caption{Model accuracy by question type across 14 models over the full dataset. All model results are scored by machine evaluation.}
    \label{tab:experiments_machine}
\end{table*}

\newpage

\end{document}